\documentclass[]{style/iromlab}

\usepackage{amsmath,amsfonts,bm}

\def\eqref#1{equation~\ref{#1}}

\def\1{\bm{1}}

\DeclareMathAlphabet{\mathsfit}{\encodingdefault}{\sfdefault}{m}{sl}
\SetMathAlphabet{\mathsfit}{bold}{\encodingdefault}{\sfdefault}{bx}{n}

\usepackage{graphicx}                          
\graphicspath{{figures/}}                      
\usepackage{wrapfig}                           
\usepackage{amsmath,amssymb,amsfonts,bm}       
\usepackage{mathtools}
\usepackage{booktabs,multirow}                 
\usepackage[caption=false,font=footnotesize]{subfig}  
\usepackage{xcolor}                            
\usepackage[most]{tcolorbox}                   
\usepackage{xspace}
\usepackage{enumitem}                            

\newcommand{\wrapfigbox}[1]{%
  \setlength{\abovecaptionskip}{4pt}%
  \vspace{-2pt}%
  \centering
  \includegraphics[width=\linewidth]{#1}%
  \par\vspace{-2pt}}

\ifdefined\FINAL
  \newcommand{\DraftOnly}[1]{}
\else
  \newcommand{\DraftOnly}[1]{#1}
\fi

\definecolor{gtcolor}{RGB}{0, 90, 200}     
\definecolor{bccolor}{RGB}{200, 30, 30}    
\definecolor{klcolor}{RGB}{0, 140, 60}     
\definecolor{lzcolor}{RGB}{150, 0, 180}    

\ifdefined\FINAL
     \newcommand{\gtc}[1]{}
     \newcommand{\bcc}[1]{}
     \newcommand{\klc}[1]{}
     \newcommand{\lzc}[1]{}
\else
  
  \newcommand{\gtc}[1]{\textcolor{gtcolor}{[GT: #1]}}
  
  \newcommand{\bcc}[1]{\textcolor{bccolor}{[BC: #1]}}
  
  \newcommand{\klc}[1]{\textcolor{klcolor}{[KL: #1]}}
  
  \newcommand{\lzc}[1]{\textcolor{lzcolor}{[LZ: #1]}}
\fi

\newcommand{\eg}{e.g.,\xspace}

\usepackage{needspace}

\renewcommand{\wrapfigbox}[1]{%
  \centering
  \includegraphics[width=\linewidth]{#1}\par}

\author[1*]{Lihan Zha}
\author[1*]{Shresth Grover}
\author[1]{Tenny Yin}
\author[1]{Samuel M. Bateman}
\author[1]{Hengkai Pan}
\author[2]{Mengchao Zhang}
\author[2]{Aykut Onol}
\author[3]{Allen Z. Ren}
\author[1,\ensuremath{\dagger}]{Dhruv Shah}
\author[1,\ensuremath{\dagger}]{Anirudha Majumdar}

\affiliation[1]{Princeton University}
\affiliation[2]{Toyota Research Institute}
\affiliation[3]{Physical Intelligence}
\contribution[*]{Equal contribution.}
\contribution[\ensuremath{\dagger}]{Equal advising.}
\website{https://ego-lap.github.io/}{https://ego-lap.github.io}

\begin{document}

\title{EgoLAP: Learning from Egocentric Human Data through Language-Action Reasoning}

\abstract{%


Egocentric human data offer a path to scaling robot learning beyond costly robot demonstrations, yet the embodiment gap makes raw human trajectories a poor supervisory target for control. 
Our key insight is that, although low-level actions are embodiment-specific, their underlying \emph{motion intent} can capture task-relevant structure that transfers across humans and robots.
We introduce \textsc{EgoLAP}, a VLA pre-training framework that jointly learns from human and robot trajectories through a shared language-based action chain-of-thought.
EgoLAP expresses motion intent as structured, temporally abstracted language actions and pairs them with motion-level reasoning grounded in scene geometry, physics, and object affordances. 
Across extensive real-world and simulated experiments, EgoLAP transfers human experience to robot control more effectively than alternative action representations and reaches $80.1\%$ mean real-world task progress, a $2.3\times$ performance gain over alternative action representations. Motion-level reasoning also outperforms a composite reasoning format that combines subtask, object-box, and visual-trace reasoning.
%
}

\maketitle

\section{INTRODUCTION}
\label{sec:introduction}
\begin{figure}[H]
  \centering
  \includegraphics[width=\columnwidth]{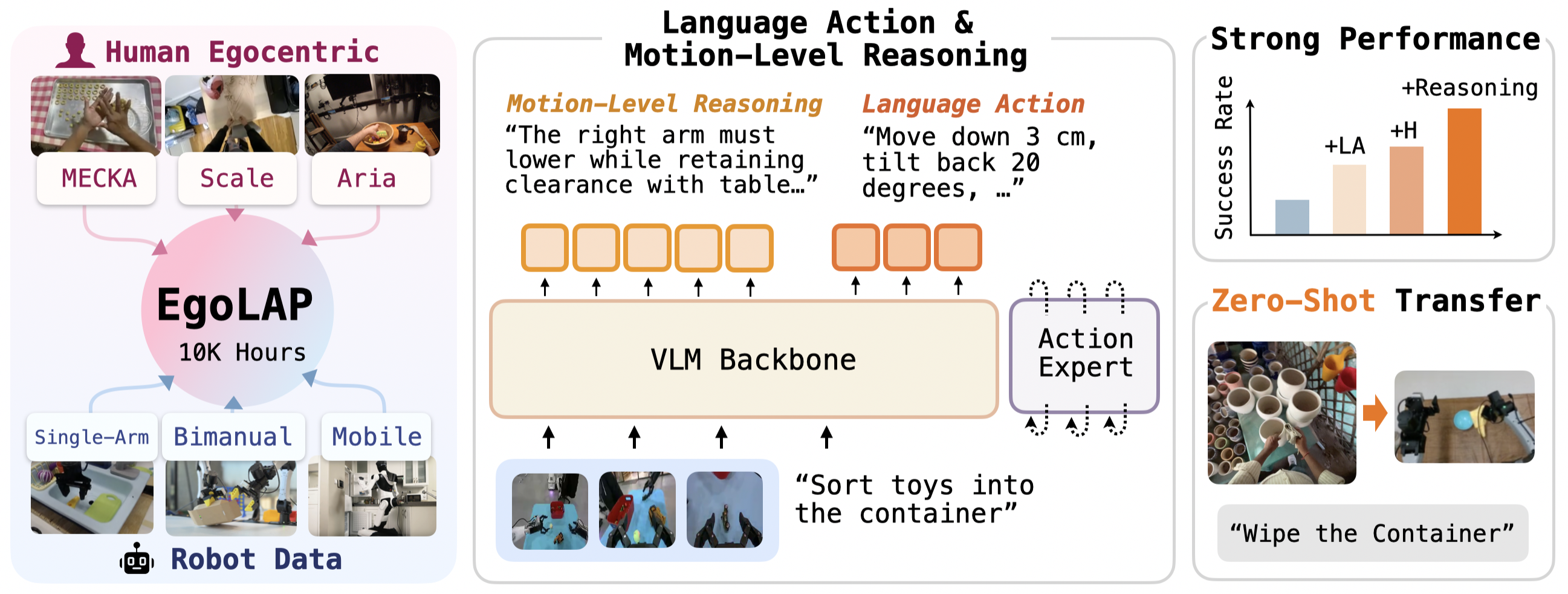}
  \vspace{-1.4em}
  \caption{\textbf{EgoLAP turns egocentric human experience into transferable robot control.}
  Human hands and robots differ in embodiment and low-level execution, yet share the same underlying motion intent when performing a task. EgoLAP captures this intent through motion-level reasoning and language actions, providing a shared supervision target across both data sources and leading to a $2.3\times$ performance gain over alternative action representations.}
  \label{fig:teaser}
\end{figure}

Robot foundation models have advanced rapidly by scaling robot demonstrations across tasks, environments, and embodiments~\citep{openx2023,khazatsky2024droid,physicalintelligence2025pi05,brohan2023rt2,kim2024openvla,ye2026worldactionmodelszeroshot}, yet such demonstrations remain costly to collect and narrow relative to the breadth of human activity. Egocentric human video offers the scale and behavioral diversity that robot data lacks, capturing long-horizon manipulation in everyday environments and growing independently of any robot platform. 

To leverage human data for robot learning, prior work extracts intermediate representations through hand-engineered pipelines, such as interaction hotspots and affordances~\citep{nagarajan2019hotspots,srirama2024hrp}, hand poses or image-space tracks~\citep{bharadhwaj2023passive,wen2024atm,bharadhwaj2024track2act,ren2025motiontracks}, latent plans~\citep{wang2023mimicplay,jain2024vid2robot,bharadhwaj2024gen2act}, or visually edited embodiments~\citep{bahl2022whirl,lepert2025phantom,lepert2026masquerade}. 
These representations capture useful facets of human behavior, but inherit the inductive biases and failure modes of their extraction pipelines that restricts scalability. 
Moreover, differences in morphology, kinematics, and control create an embodiment gap that limits the direct transfer of action-level knowledge from human data to robot policies~\citep{cai2025innonscalingegocentricmanipulation}.
Our key insight is that humans and robots need not share low-level actions; instead, they need to share \emph{motion intent}: what motion needs to be produced, and why it's appropriate given the current physical state. We represent the \emph{what} using \emph{language actions}: structured, temporally abstract descriptions of changes in pose and gripper state, such as ``move forward 3\,cm, rotate clockwise 14\,degrees, and close gripper.'' Unlike embodiment-specific trajectories or non-linguistic action tokens~\citep{pertsch2025fast,liu2026oatpolicy}, language actions reuse the vocabulary and compositional structure of a vision--language model (VLM), thereby providing control-relevant supervision within its pretrained semantic space~\citep{zha2026laplanguageactionpretrainingenables,hancock2025actionslanguage,grover2025enhancinggeneralizationvisionlanguageactionmodels}. However, the same task may require different motions across embodiments, so specifying \emph{what} motion to execute is insufficient. We therefore introduce \emph{motion-level reasoning} to represent the \emph{why}: local, physically grounded rationales that relate contact, geometry, and object affordances to the intended effect of the next motion. Unlike subtask plans or visual traces, which primarily specify what action to take or where to move~\citep{zawalski2024ecot,sun2025emmax,sun2026ervla}, motion-level reasoning captures the physical relationship between the current state and the motion required to achieve the intended effect, providing a form of supervision that can connect directly to low-level actions, yet transfer across embodiments. Together, motion-level reasoning and language actions define a physically grounded ``action chain-of-thought''.

\noindent \textbf{Statement of contributions.} We introduce \textsc{EgoLAP}, a VLA pre-training framework that co-trains on egocentric human and robot trajectories through motion-level reasoning and language actions, achieving $44.9$ points of real-world performance more than the strongest action representation. Unlike previous embodied reasoning methods~\citep{zawalski2024ecot,sun2025emmax}, EgoLAP only runs a lightweight action expert during inference time, where the textual targets provide auxiliary supervision and require no autoregressive generation.
Across extensive zero-shot real-world and simulation experiments on a bimanual robot, language actions transfer human experience better than non-linguistic action representations~\citep{pertsch2025fast,liu2026oatpolicy}, and motion-level reasoning contributes $44.7$ points of real-world task progress more to EgoLAP than the baselines. Motion-level reasoning also strongly outperforms a strong composite of subtask, object-box, and visual-trace reasoning supervision. Finally, EgoLAP adapts effectively through reinforcement learning over language actions and motion-level reasoning.

\section{RELATED WORK}
\label{sec:related_work}

\noindent \textbf{Learning Robot Policies from Egocentric Human Data. }
Human data provides scalable supervision for robot learning, but leveraging them requires bridging the embodiment gap. Prior methods rely on method-specific representations: interaction hotspots and affordances~\citep{nagarajan2019hotspots,srirama2024hrp}, hand poses or image-space tracks~\citep{bharadhwaj2023passive,wen2024atm,bharadhwaj2024track2act,ren2025motiontracks}, latent plans~\citep{wang2023mimicplay,jain2024vid2robot,bharadhwaj2024gen2act}, or visually edited embodiments~\citep{bahl2022whirl,lepert2025phantom,lepert2026masquerade}. Although effective, these hand-designed interfaces require modular processing (e.g., hand-pose estimation, point tracking, or embodiment editing) and inherit its inductive biases and errors, limiting their scalability. The gap is evident in direct policy learning: despite the 3,000 hours of egocentric video available in Ego4D~\citep{grauman2022ego4d}, MimicPlay and Motion Tracks train on only about 10 minutes of task-specific human video, ATM evaluates transfer from 100 human videos, and Phantom trains on 250--350 demonstrations per task~\citep{wang2023mimicplay,ren2025motiontracks,wen2024atm,lepert2025phantom}.

Another line directly trains policies on large-scale egocentric human datasets~\citep{punamiya2026egoverseegocentrichumandataset,hoque2026egodexlearningdexterousmanipulation} along with robot data~\citep{kareer2024egomimic,zheng2026egoscale,punamiya2025egobridge,li2026egowamworldactionmodels,qiu2025humanoidpolicy,kim2026egopi,yang2025egovla,luo2026beingh05}. These systems typically retain embodiment-specific pose targets or encode trajectories with non-linguistic action tokens as training targets~\citep{pertsch2025fast,kareer2025emergencehumanrobottransfer,wei2026psi0openfoundationmodel}, which impede cross-embodiment transfer~\citep{zha2026laplanguageactionpretrainingenables}. Language-Action Pre-training (LAP)~\citep{zha2026laplanguageactionpretrainingenables} instead expresses motion as structured natural language, showing improved zero-shot transfer among robots. EgoLAP extends this representation to human--robot co-training and pairs it with motion-level reasoning, which provides both embodiments with shared language-based targets and improves human-to-robot transfer.

\noindent \textbf{Action Representations for VLAs. }
VLAs commonly supervise a VLM with discretized action tokens~\citep{brohan2023rt2,kim2024openvla}, sometimes alongside a separate continuous action expert~\citep{physicalintelligence2025pi05}. The discrete target is especially consequential when gradients from the action expert do not reach the VLM backbone~\citep{driess2025knowledgeinsulation}. Learned tokenizers such as FAST, OAT, ActionCodec, and X-Tokenizer optimize action compression or token structure~\citep{pertsch2025fast,liu2026oatpolicy,dong2026actioncodec,kang2026xtokenizermultimodalactiontokenizer}; LAP instead describes motion in natural language to remain compatible with the VLM's pretrained semantic space~\citep{zha2026laplanguageactionpretrainingenables}. EgoLAP applies this representation to bimanual human--robot co-training and shows that motion-level reasoning yields larger gains with language actions than with non-linguistic action tokens.

\noindent \textbf{Reasoning for Visuomotor Control. }
Prior work augments visuomotor policies with textual reasoning over goals, subtask plans, and spatial cues, or with visual intermediates such as affordances and 2D point traces~\citep{zawalski2024ecot,sun2025emmax,zhao2025cotvla,lee2025molmoact,yuan2026embodiedr1,chen2025ecotlite,sun2026ervla}. With human video, plans, reasoning labels, and dense annotations improve policy performance under distribution shift~\citep{mu2023embodiedgpt,fu2025metis,kim2026demian,clark2025rad}. These results establish the value of reasoning supervision, but leave unresolved which reasoning content best supports cross-embodiment motion transfer. For example, DeMiAn finds that no single annotation type performs best across tasks~\citep{kim2026demian}, while a large-scale study finds no benefit from conditioning actions on chain-of-thought traces~\citep{lin2026systematiccotraining}.

This gap motivates \emph{motion-level reasoning}: supervision that captures the embodiment-invariant physical rationale underlying human and robot motions, thereby bridging the two data sources. Unlike prior approaches~\citep{zawalski2024ecot,chen2025ecotlite}, EgoLAP requires neither multi-stage distillation nor text generation at inference, simplifying training and ensuring fast inference.

\section{METHOD}
\label{sec:method}

\begin{figure*}[t]
  \centering
  \includegraphics[width=0.98\textwidth]{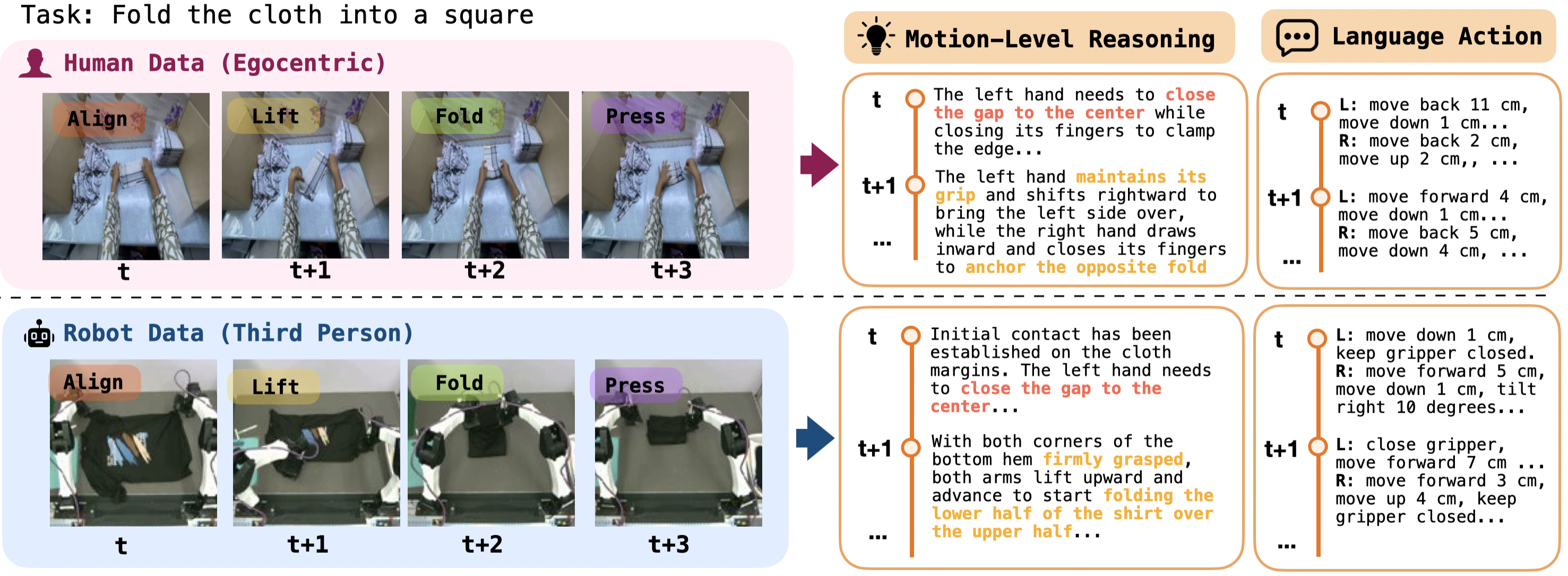}
  \caption{\textbf{EgoLAP places human and robot motion in a shared, physically grounded language space.} By capturing the physical relationship between the current state and the motion needed to produce the intended effect, motion-level reasoning provides supervision that connects directly to low-level actions that transfers across embodiments.}
  \label{fig:method}
\end{figure*}

EgoLAP is a VLA pre-training framework for transferring motion knowledge from large-scale egocentric human experience to robot manipulation. Its central design choice is to map human-hand and robot-end-effector trajectories to the same language-based interface. We first formalize this interface (Sec.~\ref{sec:problem_formulation}) and describe the model architecture and training objective (Sec.~\ref{sec:model_arc}). We then detail the construction of language actions (Sec.~\ref{sec:shared_motion}) and motion-level reasoning targets (Sec.~\ref{sec:motion_reasoning}).

\subsection{Problem Formulation}
\label{sec:problem_formulation}

Let $\mathcal{D}_{R}$ contain robot samples $(o_t,\ell,s_t,a_{t:t+K})$, where $o_t$ contains one or more RGB observations, $\ell$ is the task instruction, $s_t$ is the proprioceptive state, and $a_{t:t+K}$ is an action chunk of horizon $K$. Let $\mathcal{D}_{H}$ contain egocentric human samples $(o_t,\ell,q_{t:t+K})$, where $q$ denotes one or two tracked hand poses in the head-camera frame. For human samples, the current pose $q_t$ supplies the state input $s_t$, and the relative motion between successive poses defines the continuous action chunk $a_{t:t+K}$.

Each sample provides up to three training signals. First, a deterministic canonicalizer $G_e$, specialized only by embodiment metadata $e$ --- single arm robot, bimanual robot, or human data, produces a shared discrete action-token target $m_t$:
\begin{equation}
    m_t = G_e\!\left(q_{t:t+K}\right)
    \quad\text{or}\quad
    m_t = G_e\!\left(a_{t:t+K},s_t\right).
    \label{eq:canonicalizer}
\end{equation}
In EgoLAP, $m_t$ is a language action summarizing one second of hand or end-effector motion in an egocentric frame (Fig.~\ref{fig:method}; Sec.~\ref{sec:shared_motion}). Controlled baselines replace these discrete action tokens with FAST or OAT tokens while preserving the remaining training recipe. Second, a subset of samples includes a motion rationale $r_t$ that grounds $m_t$ in visible spatial relations, contact evidence, object affordances, and the intended physical effect of the motion (Fig.~\ref{fig:method}; Sec.~\ref{sec:motion_reasoning}). Third, the continuous action chunk $a_{t:t+K}$ directly supervises real-time control.

The model consists of a vision--language backbone $f_\theta$ and a continuous action expert $v_\phi$. The backbone predicts $m_t$ and, when available, $r_t$ on both human and robot data, thereby aligning their motion semantics. The action expert learns $a_{t:t+K}$ through flow matching. We stop action-loss gradients at the backbone boundary~\citep{driess2025knowledgeinsulation}, insulating the language-pretrained representation from the randomly initialized expert. At inference, only the action expert is rolled out to generate control commands; neither a language action nor a rationale is decoded.

\subsection{Model Architecture}
\label{sec:model_arc}

\begin{wrapfigure}{r}{0.42\textwidth}
  \vspace{-8pt}
  \wrapfigbox{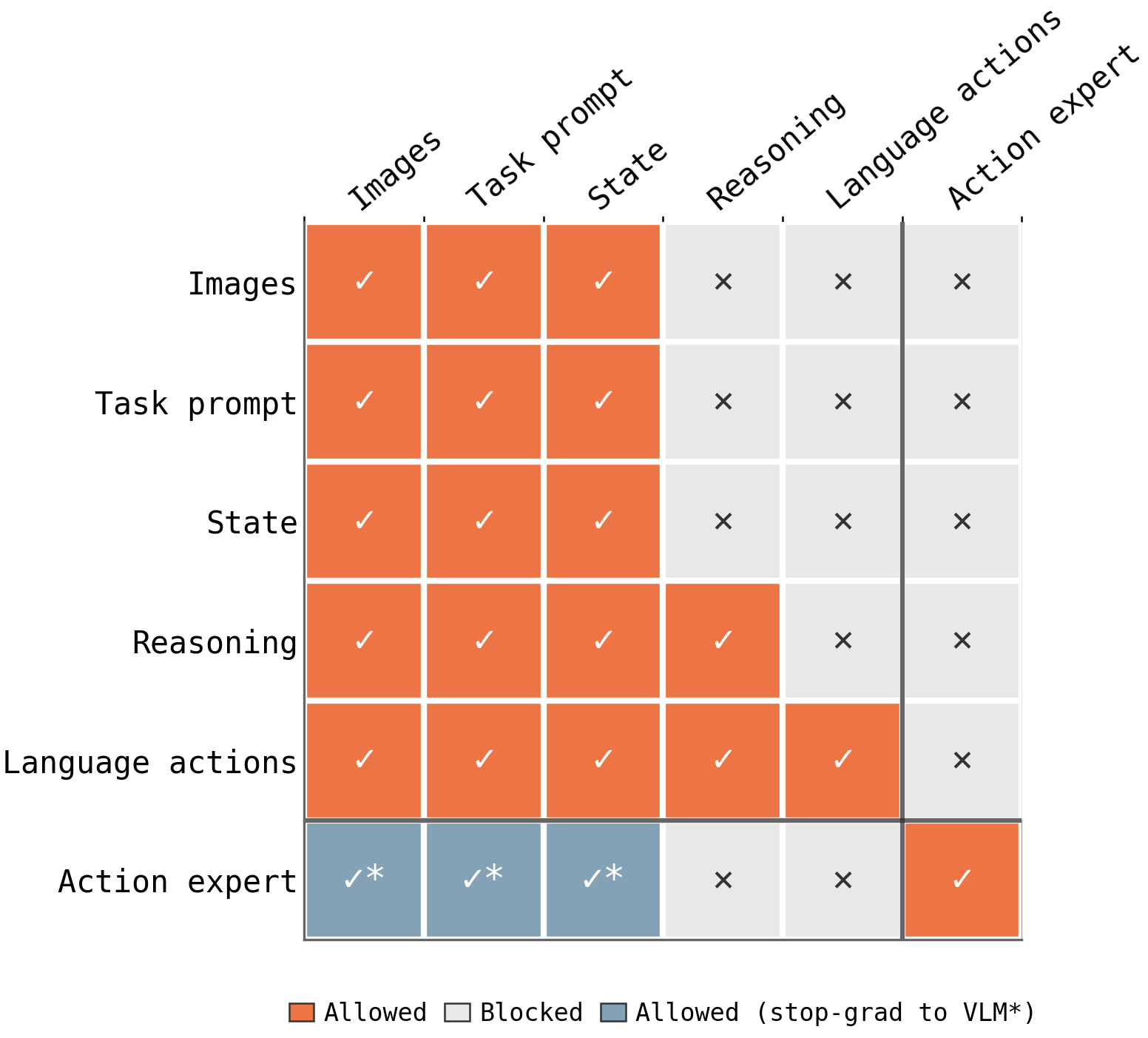}
  \caption{\textbf{Attention and gradient structure.} Orange marks allowed attention, gray blocked attention, and blue attention whose gradients are stopped at the VLM boundary. Reasoning and language action tokens are causally masked and condition on the observation prefix; action-expert tokens never attend to them, so control requires no text generation.}
  \label{fig:attention_mask}
\end{wrapfigure}

EgoLAP uses a Mixture-of-Transformers (MoT) architecture~\citep{liang2024mixture}, pairing a PaliGemma-2B vision--language backbone~\citep{beyer2024paligemma} with a 300M-parameter action expert initialized from scratch. Appendix~\ref{app:model_architecture} provides the full architecture diagram and tensor shapes.

\noindent\textbf{Inputs.}
The VLM receives up to three RGB images---one base-camera view and two wrist-camera views---together with the task instruction and proprioceptive state. Missing camera views are zero-padded. Following $\pi_{0.5}$, the normalized state is discretized into 256 uniform bins and appended to the prompt as text. For our Cartesian policies, the state encodes end-effector position, continuous 6D rotation~\citep{zhou2019continuity}, and gripper state.

\noindent\textbf{Attention and gradient structure.}
Figure~\ref{fig:attention_mask} shows the structured attention mask used by the policy. Image, instruction, and state tokens form a bidirectional prefix. Reasoning and language action tokens attend to this prefix and are casual within themselves. The action expert attends to its own tokens and to the observation--instruction--state prefix, but it cannot attend to either reasoning or language action tokens. Consequently, during inference time, no text generation is required. Knowledge insulation stops gradients from the action expert before they enter the VLM~\citep{driess2025knowledgeinsulation}; the backbone is therefore only optimized by the textual targets, while the action expert learns to consume its representations.

\noindent\textbf{Training objective.}
Let $D$ denote the action dimension. The action expert, parameterized by flow matching~\citep{physicalintelligence2025pi05}, predicts $a_{t:t+K}\in\mathbb{R}^{K\times D}$ in delta end-effector pose space, and is optimized by a flow matching loss $\mathcal{L}_{\mathrm{FM}}$. Let $\mathcal{L}_{\mathrm{text}}$ denote the expected masked token-level objective for language action and motion-level reasoning, and let $\lambda_{\mathrm{text}}$ weight the two objectives. The complete training loss is
\begin{equation}
    \mathcal{L}
    =\mathbb{E}_{(x_t,a_{t:t+K},m_t,r_t)\sim p_{\mathrm{mix}}}
      \left[\lambda_{\mathrm{text}}\mathcal{L}_{\mathrm{text}}
      +\mathcal{L}_{\mathrm{FM}}\right]
    \label{eq:full_objective}
\end{equation}
where $p_{\mathrm{mix}}$ is the fixed human--robot sampling distribution. In our model, $K{=}30$ is the action horizon. We set $\lambda_{\mathrm{text}}{=}0.8$, slightly downweighting the autoregressive VLM objective relative to continuous flow matching to align their convergence speed.

At inference, the action expert starts from Gaussian noise at $\tau{=}1$ and integrates its learned velocity field to $\tau{=}0$, conditioned only on the current observation, instruction, and state. Because the textual targets are absent from the expert's attention context, EgoLAP retains real-time continuous control without test-time language generation. Appendix~\ref{app:flow_matching} gives the complete flow-matching and inference specification, while Appendix~\ref{app:stochastic_supervision} defines the cross-entropy loss $\mathcal{L}_{\mathrm{text}}$.

\subsection{Language Actions}
\label{sec:shared_motion}

Following LAP~\citep{zha2026laplanguageactionpretrainingenables}, we supervise the VLM backbone with language actions: deterministic natural-language descriptions of hand or end-effector motion over a fixed horizon. Any trajectory with known Cartesian poses can be converted automatically; this conversion requires no manual labeling, VLM labeling, or learned action tokenizer.

\noindent\textbf{Egocentric canonicalization.}
Raw trajectories cannot share a representation unless their coordinate conventions are aligned. For each sample, we define an egocentric action frame $F_t=(R_t^F,p_t^F)$ at the start of the prediction window. We use the end-effector frame for single-arm robots with a wrist camera, the base frame for bimanual robots and single-arm robots without a wrist camera, and the head-mounted camera frame for egocentric human data. Each action frame is thus approximately aligned with its corresponding egocentric image frame—the wrist camera for single-arm robots, the base camera for bimanual robots, and the head-mounted camera for humans. Because these camera-centric frames follow broadly similar viewing conventions, motion directions are more consistent across embodiments than they would be in embodiment-specific world coordinates.

For an end-effector or hand pose $(R_\tau,p_\tau)$, we express the pose in this frame as
\begin{equation}
    \bar p_\tau=(R_t^F)^\top(p_\tau-p_t^F),
    \qquad
    \bar R_\tau=(R_t^F)^\top R_\tau .
    \label{eq:egocentric_transform}
\end{equation}
We compute the net translation $\Delta p=\bar p_{t+1\mathrm{s}}-\bar p_t$ and relative rotation $\Delta R=\bar R_t^\top\bar R_{t+1\mathrm{s}}$ over a one-second window. For bimanual samples, we apply the same procedure independently to the left and right trajectories and retain arm identity. Rotations are converted to extrinsic XYZ Euler angles before serialization.

\noindent\textbf{Deterministic language actions.}
We serialize $(\Delta p,\Delta R,\Delta c)$---where $\Delta c$ denotes a gripper or hand-state transition---using the structured grammar from LAP:
\begin{equation}
m_t^{\mathrm{lang}}
=
\text{``}\langle\text{verb}\rangle\;
\langle\text{direction}\rangle\;
\langle\text{magnitude}\rangle\;
\langle\text{unit}\rangle\text{''}.
\end{equation}
Here, \emph{move} denotes translation, \emph{tilt} denotes roll or pitch, and \emph{rotate} denotes yaw. Translational magnitudes are rounded to integer centimeters and rotational magnitudes to multiples of five degrees; components that round to zero are omitted. We append an open, close, or maintain-state phrase for the gripper or hand and use explicit \emph{left arm} and \emph{right arm} clauses for bimanual motion. For example: ``Left arm: move forward 5~cm, move left 2~cm, tilt left 10~degrees, rotate clockwise 10~degrees, close gripper. Right arm: move down 10~cm, tilt forward 15~degrees, open gripper.''

\noindent\textbf{Reinforcement learning on language actions.}
Because $G_e$ is deterministic and the grammar is machine-parseable, a language action is a \emph{verifiable} target: any sampled completion can be parsed and scored against the demonstrated motion with no learned reward model. We use this to post-train EgoLAP with GRPO~\citep{shao2024deepseekmath}, updating only the language backbone while the vision encoder and action expert stay frozen. Appendix~\ref{app:rl} gives the reward and full protocol, and results are presented in Sec.\ref{subsec:finetuning_efficiency}.

\subsection{Motion-Level Reasoning}
\label{sec:motion_reasoning}

A language action specifies \emph{what motion occurs}, but not \emph{why that motion is appropriate in the current physical state}. We therefore annotate subsets of human and robot trajectories with a motion rationale $r_t$. Given the task instruction, ordered trajectory frames, recent action history, and the ground-truth language action at the current frame, we produce a rationale that identifies a concrete end-effector--object relation, relevant contact or affordance evidence, and the immediate physical effect of the labeled motion. For example, a rationale for kneading dough might read:
\begin{quote}\small\itshape
The dough mass rests at the base of the metal bowl, with both palms directly contacting the top and sides of the dough. Kneading requires deforming the dough against the bowl surface. Both hands translate dominantly backward while tilting wrists up and opening fingers, pulling across the compliant dough toward the body before the next pressing stroke.
\end{quote}
Motion-level reasoning is deliberately local and physics-focused. It differs from the high-level task or subtask reasoning studied in prior work~\citep{zawalski2024ecot,chen2025ecotlite,sun2026ervla,sun2025emmax}: it lives on a lower hierarchy and directly connects to low-level actions, describing the underlying motion intent shared between embodiments. Appendix~\ref{app:reasoning_annotation} describes the annotation procedure, and Appendix~\ref{app:reasoning_examples} provides generated robot and human examples together with the labeling prompt. We apply reasoning dropout and reasoning loss dropout during training, which is discussed in Appendix \ref{app:stochastic_supervision}.
\section{EXPERIMENTS}
\label{sec:experiments}

We organize the evaluation around four questions: 1. Which action representation is most effective when co-training with human data? 2. Does motion-level reasoning improve transfer from human data to robots, and are motion-level reasoning and language actions mutually reinforcing? 3. How does motion-level reasoning compare to established reasoning formats? 4. Can EgoLAP be effectively adapted to downstream tasks using reinforcement learning?

\subsection{Experimental Design}
\label{subsec:experimental_design}

\begin{figure*}[t]
  \centering
  \includegraphics[width=0.98\textwidth]{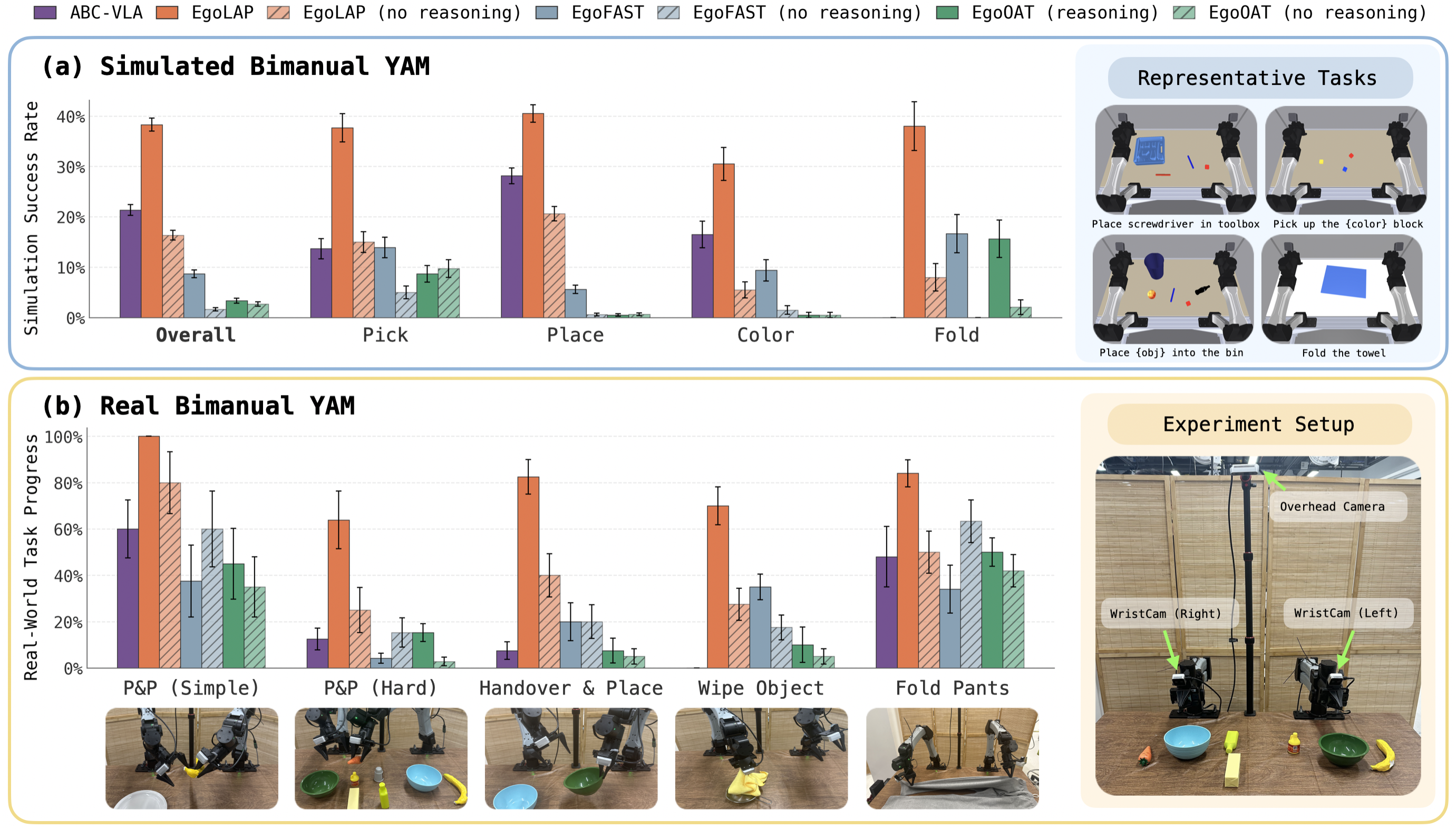}
  \caption{\textbf{Zero-shot evaluation on bimanual YAM robots.} EgoLAP achieves 80.1\% mean progress on real-world tasks—a $2.3\times$ gain over alternative action representations—and a 38.3\% success rate on simulated tasks, outperforming ABC-VLA~\citep{abc2026} by $1.8\times$. When paired with language actions, motion-level reasoning yields a 35.6-point improvement, substantially greater than the gains achieved with other action representations.}
  \label{fig:results}
\end{figure*}

\noindent\textbf{Zero-shot evaluation.}
In the real world, we evaluate our models and baselines on a customized bimanual YAM robot (Fig.~\ref{fig:results}~(b)) whose arm placement (\eg distance between the two arms) and camera configuration differ from every training embodiment. We test five behaviors spanning rigid, articulated, and deformable manipulation: Pick Place (Simple), Fold Pants, Handover \& Place, Wipe Object, and Pick Place (Complex). We run 10 trials per task, except Pick Place (Complex), for which we run 12 trials, and report mean task progress. Appendix~\ref{app:evaluation_details} describes each task and gives the complete scoring rubric. Error bar is reported as $\pm 1$ $\mathrm{SEM}$.

We complement the real-world experiments with a bimanual simulation benchmark built upon ABC~\citep{abc2026}, including 14 tasks that span picking, placing, open-ended language following, and cloth folding. Representative tasks are visualized in Fig.~\ref{fig:results}~(a).
Notably, none of our models is trained on simulation data, so this is a zero-shot simulation evaluation. We evaluate each model for 100 episodes per task, 1,400 episodes in total, and report the mean success rate for each task group. Appendix~\ref{app:evaluation_details} reports the suite composition and full evaluation protocol.

\noindent\textbf{Baselines.}
We instantiate three policy families differing only in action representation. EgoLAP uses language actions and motion-level reasoning; EgoFAST replaces language actions with FAST tokens~\citep{pertsch2025fast} and uses the same motion-level reasoning; and EgoOAT instead uses OAT tokens~\citep{liu2026oatpolicy} plus motion-level reasoning. We re-train the OAT tokenizer on our data mixture. In addition, we compare to LAP~\citep{zha2026laplanguageactionpretrainingenables}, which is EgoLAP trained without reasoning and without human data, as shown in Fig.~\ref{fig:robot_only}. In Sec.\ref{subsec:reasoning_formats}, we compare EgoLAP with our reimplemented ECoT baselines, which replaces motion-level reasoning with subtask plan, bounding box and visual trace reasoning.
Appendix~\ref{app:baseline} discusses the implementation details for baselines.

Unless stated otherwise, all variants share everything except the action representation --- same backbone, same architecture, same data mixture, same data pre-processing (\eg egocentric canonicalization in Sec.~\ref{sec:shared_motion}), same optimization budget, and same evaluation protocol. All action representations correspond to the same one-second action chunk. We train every variant for 40k steps and select its checkpoint with the same validation procedure before the final evaluation; the only exception is EgoLAP, which converges faster and we train it up to 30k steps (Appendix~\ref{app:additional_results}). Appendix~\ref{app:training_details} reports the shared training recipe and data mixture.
Notably, within our matched implementation, EgoFAST reproduces the discrete action-token interface of $\pi_{0.5}$~\citep{physicalintelligence2025pi05}, differing primarily in its training mixture. We also compare with the released checkpoint for ABC-VLA~\citep{abc2026}, which is trained directly on the ABC YAM platform.

No publicly released multi-task world-action model supports the bimanual YAM platform used in our evaluation, so we do not include WAMs in the zero-shot comparison.

\FloatBarrier
\Needspace{23\baselineskip}
\subsection{Language Actions Transfer Human Experience Most Effectively}
\label{subsec:representation_transfer}

\begin{wrapfigure}{r}{0.30\textwidth}
  \wrapfigbox{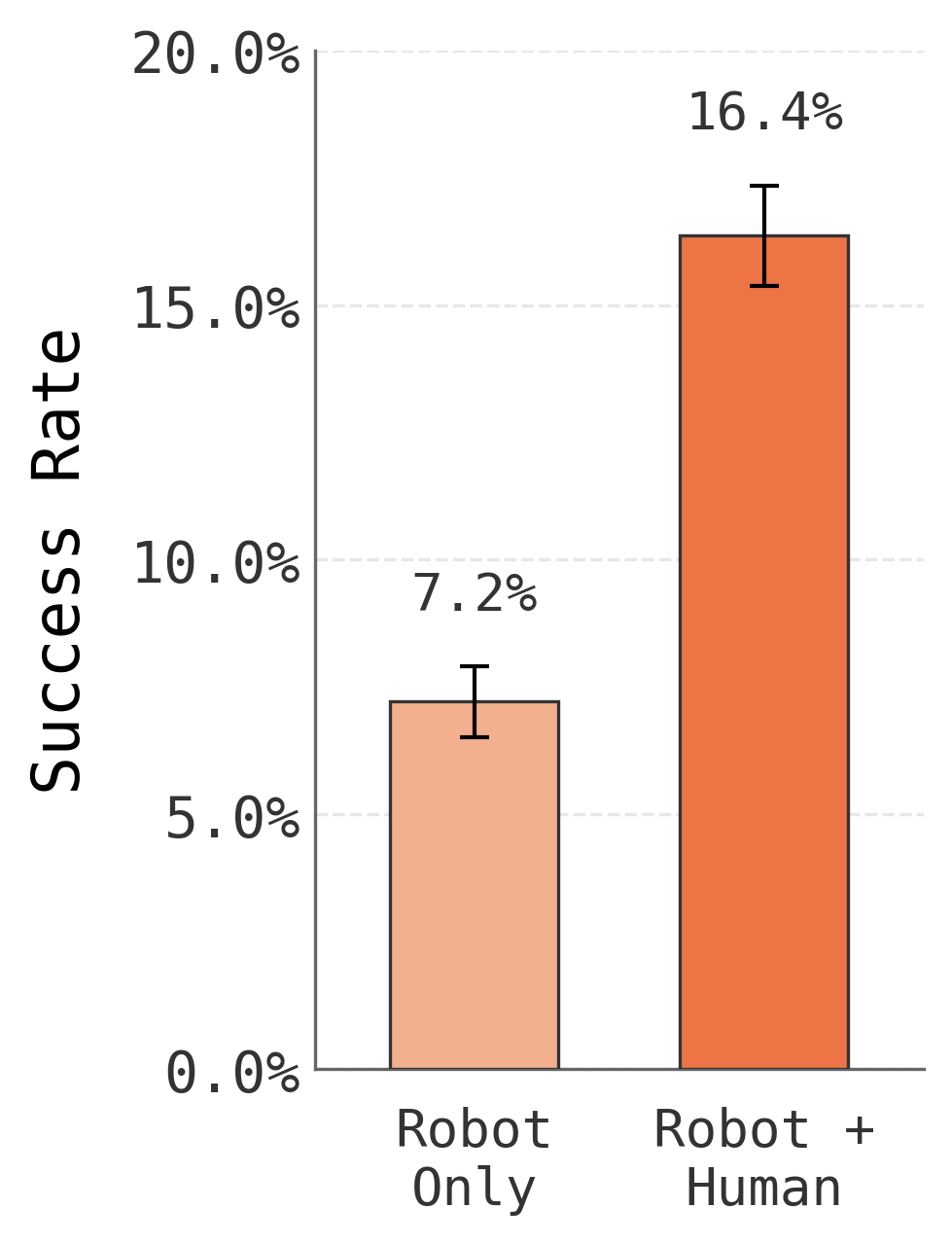}
  \caption{\textbf{Human data helps with EgoLAP.} Adding egocentric human data more than doubles simulation success for the same language action policy.}
  \label{fig:robot_only}
\end{wrapfigure}

We first show that EgoLAP benefits from human data. Holding the representation fixed, we compare EgoLAP (no reasoning) trained on human plus robot data against the same policy trained on robot data alone, i.e., LAP~\cite{zha2026laplanguageactionpretrainingenables} trained on our data mixture. Adding human data lifts simulation success from $7.2\%$ to $16.4\%$, a $2.3\times$ gain (Fig.~\ref{fig:robot_only}) The simulation suite is the harder test of transfer, since neither policy has ever seen the simulator, and it is there that human data pays off most. Appendix~\ref{app:baseline} specifies the matched robot-only training setup.

We next compare action representations. Changing only the action representation and using no reasoning, EgoLAP reaches $16.4\%$ success in simulation and $44.5\%$ progress in the real world, against $1.6\%$ and $35.2\%$ for EgoFAST and $2.7\%$ and $18.0\%$ for EgoOAT. The margin is largest exactly where the domain gap is biggest: on the unseen simulator, language actions are an order of magnitude better than FAST and OAT. This strongly suggests that language actions transfer human experience to robot tasks better than non-linguistic action tokenizers do, even without any reasoning supervision. We believe this is because language actions align with the VLM's pre-training distribution, so human and robot motions that share an intent are described by the same words and land in the same region of a representation space the backbone already understands. The full EgoLAP policy also beats ABC-VLA, which is trained directly on the data from the same simulation environment, by $16.9$ points in simulation and $54.5$ points in the real world. This strongly suggests EgoLAP overcomes an in-domain data advantage and generalizes better by transferring human data knowledge.

\subsection{Motion-Level Reasoning and Language Actions Are Mutually Reinforcing}
\label{subsec:reasoning_synergy}

We further compare each EgoX (no reasoning) policy with its reasoning-supervised counterpart. Motion-level reasoning raises EgoLAP from $44.5\%$ to $80.1\%$ mean progress in the real world, a $35.6$-point gain that holds on every one of the five tasks (Fig.~\ref{fig:results}~(b)), and from $16.4\%$ to $38.3\%$ in simulation, a $21.9$-point gain that holds on every task family (Fig.~\ref{fig:results}~(a)). This first suggests that motion-level reasoning is very helpful on its own. The same supervision, however, adds only $7.1$ points to EgoFAST in simulation and in fact \emph{hurts} it in the real world, dropping it from $35.2\%$ to $26.1\%$. EgoOAT benefits more modestly, gaining $0.7$ points in simulation and $7.6$ points in the real world.

This gap strongly suggests a synergy between language actions and motion-level reasoning. As shown in Fig.~\ref{fig:method}, the two form a single textual ``action chain-of-thought'', moving from reasoning about \emph{why} a motion is appropriate, grounded in visuals and physics, to the actual motion to execute. Non-linguistic action tokens like FAST and OAT lack this semantic correspondence, leaving the backbone to infer the correlation from co-occurrence alone, without transferring the knowledge already in the VLM backbone.

\Needspace{15\baselineskip}
\begin{wraptable}{r}{0.42\textwidth}
  \centering
  \small
  \caption{\textbf{Human data strengthens reasoning.} Zero-shot success (\%) on the 14-task simulation suite.}
  \label{tab:robot_only_reasoning}
  \setlength{\tabcolsep}{5pt}
  \begin{tabular}{lcc}
    \toprule
    Training data & No reasoning & Reasoning \\
    \midrule
    Robot only & 7.2 & 18.4 \\
    Robot + human & 16.4 & \textbf{38.3} \\
    \bottomrule
  \end{tabular}
\end{wraptable}

\noindent\textbf{Human data strengthens reasoning.}
To test whether robot-side rationale supervision alone explains the gain, we compare language-action policies trained with and without human data on the same 14-task simulation suite (Table~\ref{tab:robot_only_reasoning}). Reasoning improves robot-only success from $7.2\%$ to $18.4\%$ ($+11.2$ points), versus $16.4\%$ to $38.3\%$ ($+21.9$ points) with human data. Adding human data therefore yields $9.2$ points without reasoning and $19.9$ points with reasoning: robot-side rationale supervision alone does not explain the full gain.

\Needspace{24\baselineskip}
\subsection{Motion-Level Reasoning Outperforms Established Reasoning Formats}
\label{subsec:reasoning_formats}

\begin{wrapfigure}{r}{0.4\textwidth}
  \wrapfigbox{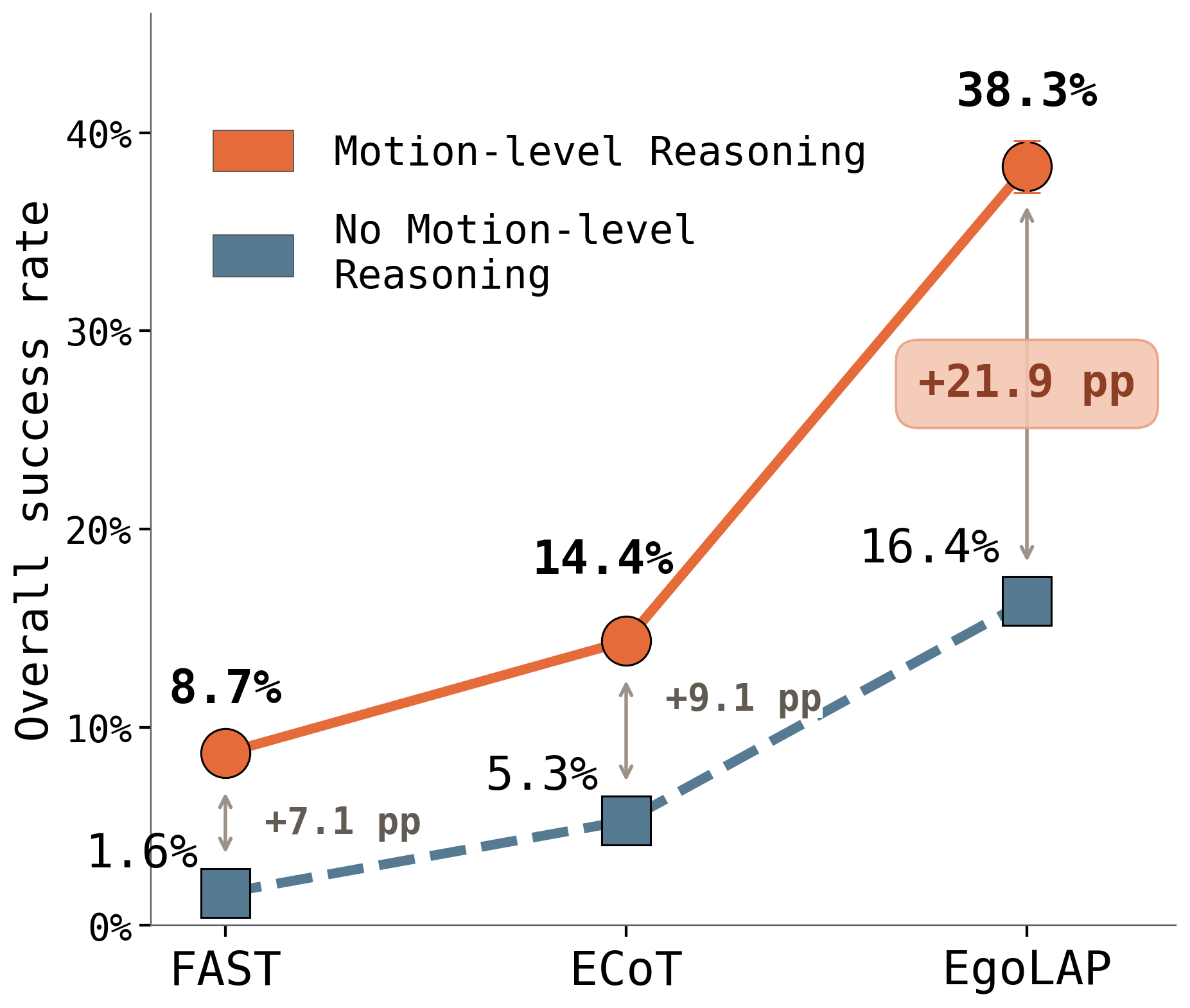}
  \caption{\textbf{Motion-level reasoning is what carries the gain.} Motion-level reasoning outperforms ECoT-style reasoning~\citep{zawalski2024ecot}, and augments it by a large margin.}
  \label{fig:ecot}
\end{wrapfigure}

We next ask whether the benefit comes from motion-level reasoning specifically or from reasoning tokens in general. We compare against a strong composite target in the style of ECoT~\citep{zawalski2024ecot}, which predicts the current subtask, visible-object bounding boxes, and a visual trace before the language action. Motion-level reasoning helps every target we tried (Fig.~\ref{fig:ecot}): it adds $7.1$ points to FAST, $9.1$ points to the ECoT composite, and $21.9$ points to plain language actions. The composite itself, however, does not help. On its own it reaches $5.3\%$, well below the $16.4\%$ of language actions with no reasoning at all, and stacking it in front of motion-level reasoning caps the policy at $14.4\%$, less than half of the $38.3\%$ obtained by motion-level reasoning alone.

Because all action-representation variants are trained on the same trajectories with identical action-conditioned rationales, and the ECoT-style control retains the same language-action target while adding subtask, box, and future-trace supervision, EgoLAP’s substantially larger gain cannot be attributed to privileged action information or merely to additional auxiliary supervision; it specifically demonstrates the effectiveness of pairing motion-level rationales with a language-based action interface.

\Needspace{25\baselineskip}
\subsection{Adapting EgoLAP to Downstream Tasks}
\label{subsec:finetuning_efficiency}
\begin{wrapfigure}{r}{0.32\textwidth}
  \wrapfigbox{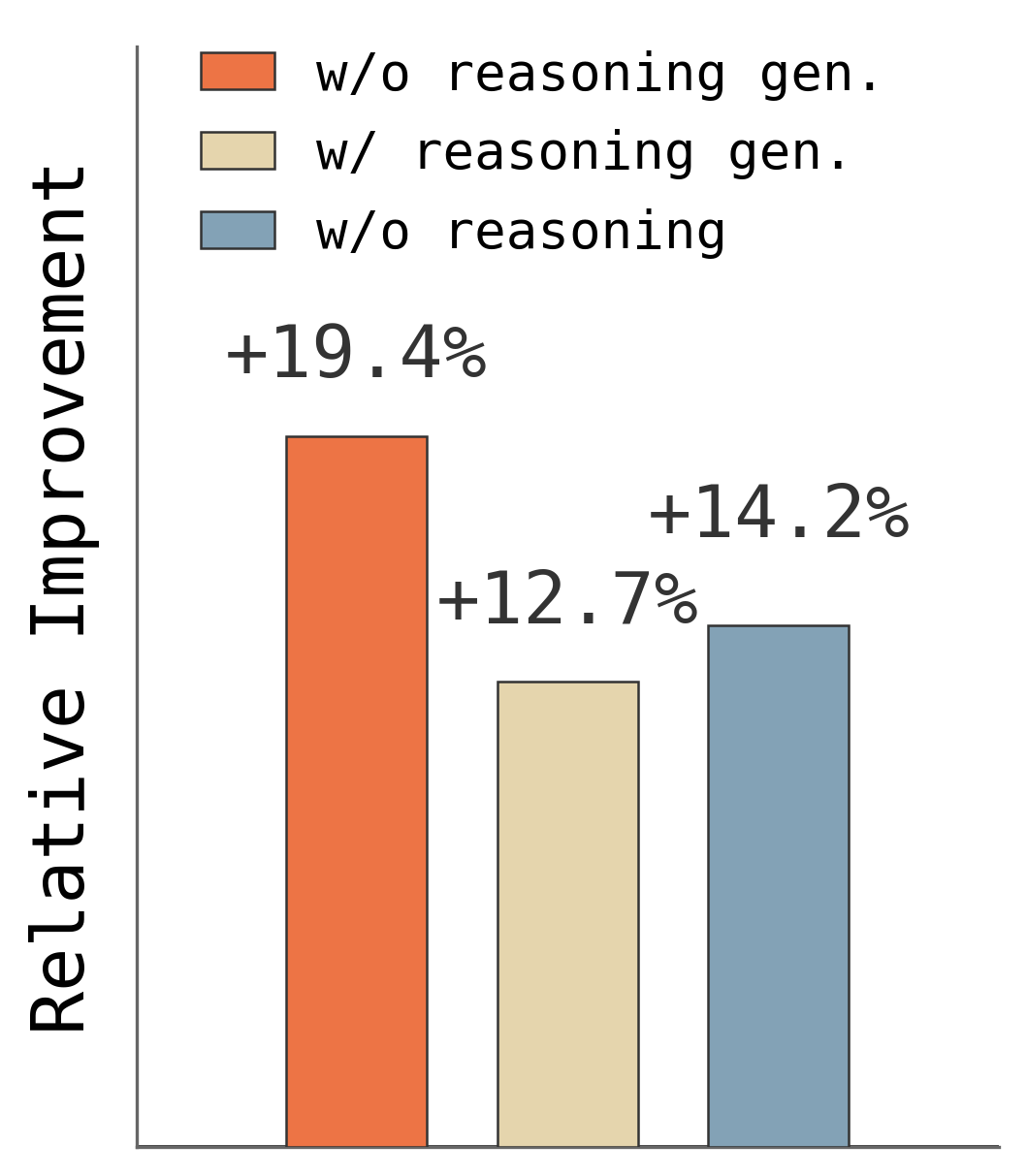}
  \caption{GRPO over language actions over simulation tasks.}
  \label{fig:rl}
\end{wrapfigure}


Following Sec.~\ref{sec:motion_reasoning}, we post-train EgoLAP with GRPO~\citep{shao2024deepseekmath} over the language actions on ABC~\citep{abc2026}, keeping the vision and action expert frozen.
We construct the language action reward by summing the negative squared translation, geodesic rotation, and a binary gripper error of the parsed language action.
This can be seen as an indirect behavior cloning alignment reward of the language actions.
We test three variations: reasoning pre-training without reasoning generation (i.e., only generates language actions during sampling), reasoning pre-training with reasoning generation, and the non-reasoning baseline (w/o reasoning).
Appendix~\ref{app:rl} gives the reward, objective, and full protocol.

Even with a frozen expert, RL improves simulation success in all three arms (Fig.~\ref{fig:rl}). Comparing the orange bar and the blue bar suggests reasoning pre-training supplies a stronger initialization for a stronger post-RL policy. Secondly, generating reasoning during RL sampling \emph{hurts} performance relative to only sampling language actions ($+12.7\%$ vs.\ $+19.4\%$): ABC rationales were never seen in training, so the sampled rationales are out of distribution, which makes the subsequent language actions generation also worse. We leave investigating how to better leverage motion-level reasoning for RL in future works.

\Needspace{19\baselineskip}
\subsection{Reasoning Aligns Skill Structure Across Embodiments}
\label{subsec:skill_alignment}

\begin{wraptable}{r}{0.50\textwidth}
  \centering
  \small
  \caption{\textbf{Motion-level reasoning improves cross-embodiment skill retrieval.} All values are percentages.}
  \vspace{5pt}
  \label{tab:cross_embodiment_alignment}
  \setlength{\tabcolsep}{5pt}
  \begin{tabular}{@{}cccc@{}}
\toprule
Retrieval & w/o & EgoLAP & $\Delta$ \\
\midrule
R$\rightarrow$H $\cdot$ 3w R@1 & 51 & \textbf{76} & $\mathbf{+25}$ \\
H$\rightarrow$R $\cdot$ 3w R@1 & 42 & \textbf{63} & $\mathbf{+21}$ \\
R$\rightarrow$H $\cdot$ 2w R@1 & 69 & \textbf{92} & $\mathbf{+23}$ \\
R$\rightarrow$H $\cdot$ 3w MRR & 61 & \textbf{80} & $\mathbf{+19}$ \\
\bottomrule
\end{tabular}

\end{wraptable}

We test matched checkpoints differing only in reasoning supervision. We mean-pool final-layer features from $1{,}002$ balanced human and robot frames, subtract each embodiment mean, and perform cross-embodiment cosine retrieval. In Tab.~\ref{tab:cross_embodiment_alignment}, R$\rightarrow$H means a robot query ranks human frames; H$\rightarrow$R is the reverse. ``3w'' and ``2w'' denote three- and two-skill splits; R@1 is the fraction whose nearest frame has the same skill; MRR is the mean reciprocal rank of the first same-skill match; ``w/o'' is the matched checkpoint without reasoning; and $\Delta$ is the absolute gain from reasoning in percentage points. We report paired task-bootstrap confidence intervals.

Robot-to-human R@1 rises from 51\% to 76\% (+25 points, 95\% CI $[16,39]$, $p<.005$), while the reverse direction rises from 42\% to 63\% (+21 points, $p=.04$). The two-skill split improves from 69\% to 92\% (+23 points, 95\% CI $[16,30]$), and MRR from 61\% to 80\% (Tab.~\ref{tab:cross_embodiment_alignment}). Because the embodiments remain linearly separable, reasoning preserves domain identity but gives both domains the same internal organization---fold near fold, place near place.

\section{CONCLUSION}

We introduce \textsc{EgoLAP}, a VLA pre-training framework that transfers egocentric human experience to robots through a shared language of motion. By pairing language actions that describe \emph{what} motion to produce with physically grounded reasoning that explains \emph{why}, EgoLAP reaches $80.1\%$ real-world task progress and $38.3\%$ simulation success on bimanual manipulation, substantially outperforming alternative action representations and reasoning formats. Beyond policy performance, motion-level reasoning improves cross-embodiment skill retrieval by up to $25$ points, suggesting that it organizes human and robot experience by shared skill structure without erasing embodiment identity. EgoLAP also adapts effectively through reinforcement learning, while its textual targets remain training-only and add no generation cost during control.

\textbf{Limitations and future work.} Language actions currently summarize fixed one-second windows with quantized translation and rotation, which may miss the dynamics or precision required for fast, contact-rich manipulation. Their linguistic form, however, can naturally support adaptive precision: future representations could combine coarse, long-horizon descriptions with fine, high-frequency corrections, while tolerating approximate labels from pose tracking, UMI, or internet video. Our evaluation remains focused on bimanual tabletop tasks. Finally, our reinforcement-learning study is confined to simulation and optimizes a motion-matching proxy rather than task reward; future work should jointly post-train motion-level reasoning and the action expert using task-level feedback.

\section*{Acknowledgments}

Toyota Research Institute (TRI) provided funds to assist the authors with their research; this article solely reflects the opinions and conclusions of its authors, and not TRI nor any other Toyota entity. We thank Google’s TPU Research Cloud (TRC) for providing access to Cloud TPUs. We also thank Tim Chu and Asher J.~Hancock for their valuable feedback on this project. Samuel M. Bateman is supported by the National Science Foundation CISE Graduate Fellowships under Grant No. 2313998. The authors were partially supported by the Office of Naval Research under Grant N00014-23-1-2148, the Army Research Laboratory under DCIST CRA W911NF-17-2-0181, and Gemini Academic Program.

\section*{AI Use Statement}
We used large language models to edit and polish the manuscript. We also used Google Gemini to generate offline motion-level reasoning labels: \texttt{gemini-3.5-flash} for DROID and \texttt{gemini-3.7-flash} for MECKA. These labels serve as training targets for the reasoning-enabled policies; Appendix~\ref{app:reasoning_annotation} documents the inputs and annotation procedure, and Appendix~\ref{app:reasoning_examples} provides the prompts and representative outputs. The authors designed and checked the labeling pipeline, reviewed the AI-assisted manuscript edits, and take responsibility for all content, claims, results, and AI-assisted artifacts in this work.

\section*{Reproducibility Statement}
We provide the methodological and experimental details needed to reproduce our results. Section~\ref{sec:method} specifies the model, objectives, shared motion representation, and motion-level reasoning formulation, while Section~\ref{subsec:experimental_design} describes the matched comparison protocol. Appendix~\ref{app:training_details} documents the architecture, prompt and action serialization, flow-matching objective, reasoning-annotation pipeline, training hyperparameters, data mixture, and baseline implementations. Appendix~\ref{app:evaluation_details} specifies the evaluation tasks and number of episodes, Appendix~\ref{app:rl} gives the reinforcement-learning setup, and Appendix~\ref{app:additional_results} reports additional results and checkpoint-selection details. The complete Gemini labeling prompts and representative annotations are included in Appendix~\ref{app:reasoning_examples}.


\bibliography{references}

@misc{openx2023,
      title={Open X-Embodiment: Robotic Learning Datasets and RT-X Models}, 
      author={Embodiment Collaboration and Abby O'Neill and Abdul Rehman and Abhinav Gupta and Abhiram Maddukuri and Abhishek Gupta and Abhishek Padalkar and Abraham Lee and Acorn Pooley and Agrim Gupta and Ajay Mandlekar and Ajinkya Jain and Albert Tung and Alex Bewley and Alex Herzog and Alex Irpan and Alexander Khazatsky and Anant Rai and Anchit Gupta and Andrew Wang and Andrey Kolobov and Anikait Singh and Animesh Garg and Aniruddha Kembhavi and Annie Xie and Anthony Brohan and Antonin Raffin and Archit Sharma and Arefeh Yavary and Arhan Jain and Ashwin Balakrishna and Ayzaan Wahid and Ben Burgess-Limerick and Beomjoon Kim and Bernhard Schölkopf and Blake Wulfe and Brian Ichter and Cewu Lu and Charles Xu and Charlotte Le and Chelsea Finn and Chen Wang and Chenfeng Xu and Cheng Chi and Chenguang Huang and Christine Chan and Christopher Agia and Chuer Pan and Chuyuan Fu and Coline Devin and Danfei Xu and Daniel Morton and Danny Driess and Daphne Chen and Deepak Pathak and Dhruv Shah and Dieter Büchler and Dinesh Jayaraman and Dmitry Kalashnikov and Dorsa Sadigh and Edward Johns and Ethan Foster and Fangchen Liu and Federico Ceola and Fei Xia and Feiyu Zhao and Felipe Vieira Frujeri and Freek Stulp and Gaoyue Zhou and Gaurav S. Sukhatme and Gautam Salhotra and Ge Yan and Gilbert Feng and Giulio Schiavi and Glen Berseth and Gregory Kahn and Guangwen Yang and Guanzhi Wang and Hao Su and Hao-Shu Fang and Haochen Shi and Henghui Bao and Heni Ben Amor and Henrik I Christensen and Hiroki Furuta and Homanga Bharadhwaj and Homer Walke and Hongjie Fang and Huy Ha and Igor Mordatch and Ilija Radosavovic and Isabel Leal and Jacky Liang and Jad Abou-Chakra and Jaehyung Kim and Jaimyn Drake and Jan Peters and Jan Schneider and Jasmine Hsu and Jay Vakil and Jeannette Bohg and Jeffrey Bingham and Jeffrey Wu and Jensen Gao and Jiaheng Hu and Jiajun Wu and Jialin Wu and Jiankai Sun and Jianlan Luo and Jiayuan Gu and Jie Tan and Jihoon Oh and Jimmy Wu and Jingpei Lu and Jingyun Yang and Jitendra Malik and João Silvério and Joey Hejna and Jonathan Booher and Jonathan Tompson and Jonathan Yang and Jordi Salvador and Joseph J. Lim and Junhyek Han and Kaiyuan Wang and Kanishka Rao and Karl Pertsch and Karol Hausman and Keegan Go and Keerthana Gopalakrishnan and Ken Goldberg and Kendra Byrne and Kenneth Oslund and Kento Kawaharazuka and Kevin Black and Kevin Lin and Kevin Zhang and Kiana Ehsani and Kiran Lekkala and Kirsty Ellis and Krishan Rana and Krishnan Srinivasan and Kuan Fang and Kunal Pratap Singh and Kuo-Hao Zeng and Kyle Hatch and Kyle Hsu and Laurent Itti and Lawrence Yunliang Chen and Lerrel Pinto and Li Fei-Fei and Liam Tan and Linxi "Jim" Fan and Lionel Ott and Lisa Lee and Luca Weihs and Magnum Chen and Marion Lepert and Marius Memmel and Masayoshi Tomizuka and Masha Itkina and Mateo Guaman Castro and Max Spero and Maximilian Du and Michael Ahn and Michael C. Yip and Mingtong Zhang and Mingyu Ding and Minho Heo and Mohan Kumar Srirama and Mohit Sharma and Moo Jin Kim and Muhammad Zubair Irshad and Naoaki Kanazawa and Nicklas Hansen and Nicolas Heess and Nikhil J Joshi and Niko Suenderhauf and Ning Liu and Norman Di Palo and Nur Muhammad Mahi Shafiullah and Oier Mees and Oliver Kroemer and Osbert Bastani and Pannag R Sanketi and Patrick "Tree" Miller and Patrick Yin and Paul Wohlhart and Peng Xu and Peter David Fagan and Peter Mitrano and Pierre Sermanet and Pieter Abbeel and Priya Sundaresan and Qiuyu Chen and Quan Vuong and Rafael Rafailov and Ran Tian and Ria Doshi and Roberto Martín-Martín and Rohan Baijal and Rosario Scalise and Rose Hendrix and Roy Lin and Runjia Qian and Ruohan Zhang and Russell Mendonca and Rutav Shah and Ryan Hoque and Ryan Julian and Samuel Bustamante and Sean Kirmani and Sergey Levine and Shan Lin and Sherry Moore and Shikhar Bahl and Shivin Dass and Shubham Sonawani and Shubham Tulsiani and Shuran Song and Sichun Xu and Siddhant Haldar and Siddharth Karamcheti and Simeon Adebola and Simon Guist and Soroush Nasiriany and Stefan Schaal and Stefan Welker and Stephen Tian and Subramanian Ramamoorthy and Sudeep Dasari and Suneel Belkhale and Sungjae Park and Suraj Nair and Suvir Mirchandani and Takayuki Osa and Tanmay Gupta and Tatsuya Harada and Tatsuya Matsushima and Ted Xiao and Thomas Kollar and Tianhe Yu and Tianli Ding and Todor Davchev and Tony Z. Zhao and Travis Armstrong and Trevor Darrell and Trinity Chung and Vidhi Jain and Vikash Kumar and Vincent Vanhoucke and Vitor Guizilini and Wei Zhan and Wenxuan Zhou and Wolfram Burgard and Xi Chen and Xiangyu Chen and Xiaolong Wang and Xinghao Zhu and Xinyang Geng and Xiyuan Liu and Xu Liangwei and Xuanlin Li and Yansong Pang and Yao Lu and Yecheng Jason Ma and Yejin Kim and Yevgen Chebotar and Yifan Zhou and Yifeng Zhu and Yilin Wu and Ying Xu and Yixuan Wang and Yonatan Bisk and Yongqiang Dou and Yoonyoung Cho and Youngwoon Lee and Yuchen Cui and Yue Cao and Yueh-Hua Wu and Yujin Tang and Yuke Zhu and Yunchu Zhang and Yunfan Jiang and Yunshuang Li and Yunzhu Li and Yusuke Iwasawa and Yutaka Matsuo and Zehan Ma and Zhuo Xu and Zichen Jeff Cui and Zichen Zhang and Zipeng Fu and Zipeng Lin},
      year={2025},
      eprint={2310.08864},
      archivePrefix={arXiv},
      primaryClass={cs.RO},
      url={https://arxiv.org/abs/2310.08864}, 
}

@misc{khazatsky2024droid,
      title={DROID: A Large-Scale In-The-Wild Robot Manipulation Dataset}, 
      author={Alexander Khazatsky and Karl Pertsch and Suraj Nair and Ashwin Balakrishna and Sudeep Dasari and Siddharth Karamcheti and Soroush Nasiriany and Mohan Kumar Srirama and Lawrence Yunliang Chen and Kirsty Ellis and Peter David Fagan and Joey Hejna and Masha Itkina and Marion Lepert and Yecheng Jason Ma and Patrick Tree Miller and Jimmy Wu and Suneel Belkhale and Shivin Dass and Huy Ha and Arhan Jain and Abraham Lee and Youngwoon Lee and Marius Memmel and Sungjae Park and Ilija Radosavovic and Kaiyuan Wang and Albert Zhan and Kevin Black and Cheng Chi and Kyle Beltran Hatch and Shan Lin and Jingpei Lu and Jean Mercat and Abdul Rehman and Pannag R Sanketi and Archit Sharma and Cody Simpson and Quan Vuong and Homer Rich Walke and Blake Wulfe and Ted Xiao and Jonathan Heewon Yang and Arefeh Yavary and Tony Z. Zhao and Christopher Agia and Rohan Baijal and Mateo Guaman Castro and Daphne Chen and Qiuyu Chen and Trinity Chung and Jaimyn Drake and Ethan Paul Foster and Jensen Gao and Vitor Guizilini and David Antonio Herrera and Minho Heo and Kyle Hsu and Jiaheng Hu and Muhammad Zubair Irshad and Donovon Jackson and Charlotte Le and Yunshuang Li and Kevin Lin and Roy Lin and Zehan Ma and Abhiram Maddukuri and Suvir Mirchandani and Daniel Morton and Tony Nguyen and Abigail O'Neill and Rosario Scalise and Derick Seale and Victor Son and Stephen Tian and Emi Tran and Andrew E. Wang and Yilin Wu and Annie Xie and Jingyun Yang and Patrick Yin and Yunchu Zhang and Osbert Bastani and Glen Berseth and Jeannette Bohg and Ken Goldberg and Abhinav Gupta and Abhishek Gupta and Dinesh Jayaraman and Joseph J Lim and Jitendra Malik and Roberto Martín-Martín and Subramanian Ramamoorthy and Dorsa Sadigh and Shuran Song and Jiajun Wu and Michael C. Yip and Yuke Zhu and Thomas Kollar and Sergey Levine and Chelsea Finn},
      year={2025},
      eprint={2403.12945},
      archivePrefix={arXiv},
      primaryClass={cs.RO},
      url={https://arxiv.org/abs/2403.12945}, 
}

@misc{punamiya2026egoverseegocentrichumandataset,
      title={EgoVerse: An Egocentric Human Dataset for Robot Learning from Around the World}, 
      author={Ryan Punamiya and Simar Kareer and Zeyi Liu and Josh Citron and Ri-Zhao Qiu and Xiongyi Cai and Alexey Gavryushin and Jiaqi Chen and Davide Liconti and Lawrence Y. Zhu and Patcharapong Aphiwetsa and Baoyu Li and Aniketh Cheluva and Pranav Kuppili and Yangcen Liu and Dhruv Patel and Aidan Gao and Hye-Young Chung and Ryan Co and Renee Zbizika and Jeff Liu and Xiaomeng Xu and Haoyu Xiong and Geng Chen and Sebastiano Oliani and Wenkai Xuan and Chenyu Yang and Xi Wang and James Fort and Richard Newcombe and Josh Gao and Jason Chong and Garrett Matsuda and Aseem Doriwala and Marc Pollefeys and Robert Katzschmann and Xiaolong Wang and Shuran Song and Judy Hoffman and Danfei Xu},
      year={2026},
      eprint={2604.07607},
      archivePrefix={arXiv},
      primaryClass={cs.RO},
      url={https://arxiv.org/abs/2604.07607}, 
}

@misc{qiu2025humanoidpolicy,
      title={Humanoid Policy ~ Human Policy}, 
      author={Ri-Zhao Qiu and Shiqi Yang and Xuxin Cheng and Chaitanya Chawla and Jialong Li and Tairan He and Ge Yan and David J. Yoon and Ryan Hoque and Lars Paulsen and Ge Yang and Jian Zhang and Sha Yi and Guanya Shi and Xiaolong Wang},
      year={2025},
      eprint={2503.13441},
      archivePrefix={arXiv},
      primaryClass={cs.RO},
      url={https://arxiv.org/abs/2503.13441}, 
}

@misc{yang2025egovla,
      title={EgoVLA: Learning Vision-Language-Action Models from Egocentric Human Videos}, 
      author={Ruihan Yang and Qinxi Yu and Yecheng Wu and Rui Yan and Borui Li and An-Chieh Cheng and Xueyan Zou and Yunhao Fang and Xuxin Cheng and Ri-Zhao Qiu and Hongxu Yin and Sifei Liu and Song Han and Yao Lu and Xiaolong Wang},
      year={2025},
      eprint={2507.12440},
      archivePrefix={arXiv},
      primaryClass={cs.RO},
      url={https://arxiv.org/abs/2507.12440}, 
}

@misc{luo2026beingh05,
      title={Being-H0.5: Scaling Human-Centric Robot Learning for Cross-Embodiment Generalization}, 
      author={Hao Luo and Ye Wang and Wanpeng Zhang and Sipeng Zheng and Ziheng Xi and Chaoyi Xu and Haiweng Xu and Haoqi Yuan and Chi Zhang and Yiqing Wang and Yicheng Feng and Zongqing Lu},
      year={2026},
      eprint={2601.12993},
      archivePrefix={arXiv},
      primaryClass={cs.RO},
      url={https://arxiv.org/abs/2601.12993}, 
}

@misc{ren2025motiontracks,
      title={Motion Tracks: A Unified Representation for Human-Robot Transfer in Few-Shot Imitation Learning}, 
      author={Juntao Ren and Priya Sundaresan and Dorsa Sadigh and Sanjiban Choudhury and Jeannette Bohg},
      year={2025},
      eprint={2501.06994},
      archivePrefix={arXiv},
      primaryClass={cs.RO},
      url={https://arxiv.org/abs/2501.06994}, 
}

@inproceedings{grauman2022ego4d,
      title={Ego4D: Around the World in 3,000 Hours of Egocentric Video},
      author={Kristen Grauman and Andrew Westbury and Eugene Byrne and Zachary Chavis and Antonino Furnari and Rohit Girdhar and Jackson Hamburger and Hao Jiang and Miao Liu and Xingyu Liu and others},
      booktitle={Proceedings of the IEEE/CVF Conference on Computer Vision and Pattern Recognition},
      pages={18995--19012},
      year={2022},
      url={https://arxiv.org/abs/2110.07058},
}

@misc{nagarajan2019hotspots,
      title={Grounded Human-Object Interaction Hotspots from Video}, 
      author={Tushar Nagarajan and Christoph Feichtenhofer and Kristen Grauman},
      year={2019},
      eprint={1812.04558},
      archivePrefix={arXiv},
      primaryClass={cs.CV},
      url={https://arxiv.org/abs/1812.04558}, 
}

@misc{srirama2024hrp,
      title={HRP: Human Affordances for Robotic Pre-Training}, 
      author={Mohan Kumar Srirama and Sudeep Dasari and Shikhar Bahl and Abhinav Gupta},
      year={2024},
      eprint={2407.18911},
      archivePrefix={arXiv},
      primaryClass={cs.RO},
      url={https://arxiv.org/abs/2407.18911}, 
}

@misc{bharadhwaj2023passive,
      title={Zero-Shot Robot Manipulation from Passive Human Videos}, 
      author={Homanga Bharadhwaj and Abhinav Gupta and Shubham Tulsiani and Vikash Kumar},
      year={2023},
      eprint={2302.02011},
      archivePrefix={arXiv},
      primaryClass={cs.RO},
      url={https://arxiv.org/abs/2302.02011}, 
}

@misc{wen2024atm,
      title={Any-point Trajectory Modeling for Policy Learning}, 
      author={Chuan Wen and Xingyu Lin and John So and Kai Chen and Qi Dou and Yang Gao and Pieter Abbeel},
      year={2024},
      eprint={2401.00025},
      archivePrefix={arXiv},
      primaryClass={cs.RO},
      url={https://arxiv.org/abs/2401.00025}, 
}

@misc{bharadhwaj2024track2act,
      title={Track2Act: Predicting Point Tracks from Internet Videos enables Generalizable Robot Manipulation}, 
      author={Homanga Bharadhwaj and Roozbeh Mottaghi and Abhinav Gupta and Shubham Tulsiani},
      year={2024},
      eprint={2405.01527},
      archivePrefix={arXiv},
      primaryClass={cs.RO},
      url={https://arxiv.org/abs/2405.01527}, 
}

@misc{wang2023mimicplay,
      title={MimicPlay: Long-Horizon Imitation Learning by Watching Human Play}, 
      author={Chen Wang and Linxi Fan and Jiankai Sun and Ruohan Zhang and Li Fei-Fei and Danfei Xu and Yuke Zhu and Anima Anandkumar},
      year={2023},
      eprint={2302.12422},
      archivePrefix={arXiv},
      primaryClass={cs.RO},
      url={https://arxiv.org/abs/2302.12422}, 
}

@misc{jain2024vid2robot,
      title={Vid2Robot: End-to-end Video-conditioned Policy Learning with Cross-Attention Transformers}, 
      author={Vidhi Jain and Maria Attarian and Nikhil J Joshi and Ayzaan Wahid and Danny Driess and Quan Vuong and Pannag R Sanketi and Pierre Sermanet and Stefan Welker and Christine Chan and Igor Gilitschenski and Yonatan Bisk and Debidatta Dwibedi},
      year={2024},
      eprint={2403.12943},
      archivePrefix={arXiv},
      primaryClass={cs.RO},
      url={https://arxiv.org/abs/2403.12943}, 
}

@misc{bharadhwaj2024gen2act,
      title={Gen2Act: Human Video Generation in Novel Scenarios enables Generalizable Robot Manipulation}, 
      author={Homanga Bharadhwaj and Debidatta Dwibedi and Abhinav Gupta and Shubham Tulsiani and Carl Doersch and Ted Xiao and Dhruv Shah and Fei Xia and Dorsa Sadigh and Sean Kirmani},
      year={2024},
      eprint={2409.16283},
      archivePrefix={arXiv},
      primaryClass={cs.RO},
      url={https://arxiv.org/abs/2409.16283}, 
}

@misc{bahl2022whirl,
      title={Human-to-Robot Imitation in the Wild}, 
      author={Shikhar Bahl and Abhinav Gupta and Deepak Pathak},
      year={2022},
      eprint={2207.09450},
      archivePrefix={arXiv},
      primaryClass={cs.RO},
      url={https://arxiv.org/abs/2207.09450}, 
}

@misc{lepert2025phantom,
      title={Phantom: Training Robots Without Robots Using Only Human Videos}, 
      author={Marion Lepert and Jiaying Fang and Jeannette Bohg},
      year={2026},
      eprint={2503.00779},
      archivePrefix={arXiv},
      primaryClass={cs.RO},
      url={https://arxiv.org/abs/2503.00779}, 
}

@misc{lepert2026masquerade,
      title={Masquerade: Learning from In-the-wild Human Videos using Data-Editing}, 
      author={Marion Lepert and Jiaying Fang and Jeannette Bohg},
      year={2025},
      eprint={2508.09976},
      archivePrefix={arXiv},
      primaryClass={cs.RO},
      url={https://arxiv.org/abs/2508.09976}, 
}

@misc{pertsch2025fast,
      title={FAST: Efficient Action Tokenization for Vision-Language-Action Models}, 
      author={Karl Pertsch and Kyle Stachowicz and Brian Ichter and Danny Driess and Suraj Nair and Quan Vuong and Oier Mees and Chelsea Finn and Sergey Levine},
      year={2025},
      eprint={2501.09747},
      archivePrefix={arXiv},
      primaryClass={cs.RO},
      url={https://arxiv.org/abs/2501.09747}, 
}

@misc{kang2026xtokenizermultimodalactiontokenizer,
      title={X-Tokenizer: A Multimodal Action Tokenizer for Vision-Language-Action Pretraining}, 
      author={Miracle Kang and Lights Shi and Lucy Liang and Roy Gan and Dongxiu Liu and Pushi Zhang and Sylas Chen and Shawn Qin and Yinan Zheng and Jinliang Zheng and Hao Wang and Xianyuan Zhan and Hang Su},
      year={2026},
      eprint={2606.14752},
      archivePrefix={arXiv},
      primaryClass={cs.CV},
      url={https://arxiv.org/abs/2606.14752}, 
}

@misc{zha2026laplanguageactionpretrainingenables,
      title={LAP: Language-Action Pre-Training Enables Zero-shot Cross-Embodiment Transfer}, 
      author={Lihan Zha and Asher J. Hancock and Mingtong Zhang and Tenny Yin and Yixuan Huang and Dhruv Shah and Allen Z. Ren and Anirudha Majumdar},
      year={2026},
      eprint={2602.10556},
      archivePrefix={arXiv},
      primaryClass={cs.RO},
      url={https://arxiv.org/abs/2602.10556}, 
}

@misc{li2026egowamworldactionmodels,
      title={EgoWAM: World Action Models Beyond Pixels with In-the-Wild Egocentric Human Data}, 
      author={Baoyu Li and Xinchen Yin and Mengying Lin and Yixin Zhang and Danfei Xu},
      year={2026},
      eprint={2607.08436},
      archivePrefix={arXiv},
      primaryClass={cs.RO},
      url={https://arxiv.org/abs/2607.08436}, 
}

@misc{brohan2023rt2,
      title={RT-2: Vision-Language-Action Models Transfer Web Knowledge to Robotic Control}, 
      author={Anthony Brohan and Noah Brown and Justice Carbajal and Yevgen Chebotar and Xi Chen and Krzysztof Choromanski and Tianli Ding and Danny Driess and Avinava Dubey and Chelsea Finn and Pete Florence and Chuyuan Fu and Montse Gonzalez Arenas and Keerthana Gopalakrishnan and Kehang Han and Karol Hausman and Alexander Herzog and Jasmine Hsu and Brian Ichter and Alex Irpan and Nikhil Joshi and Ryan Julian and Dmitry Kalashnikov and Yuheng Kuang and Isabel Leal and Lisa Lee and Tsang-Wei Edward Lee and Sergey Levine and Yao Lu and Henryk Michalewski and Igor Mordatch and Karl Pertsch and Kanishka Rao and Krista Reymann and Michael Ryoo and Grecia Salazar and Pannag Sanketi and Pierre Sermanet and Jaspiar Singh and Anikait Singh and Radu Soricut and Huong Tran and Vincent Vanhoucke and Quan Vuong and Ayzaan Wahid and Stefan Welker and Paul Wohlhart and Jialin Wu and Fei Xia and Ted Xiao and Peng Xu and Sichun Xu and Tianhe Yu and Brianna Zitkovich},
      year={2023},
      eprint={2307.15818},
      archivePrefix={arXiv},
      primaryClass={cs.RO},
      url={https://arxiv.org/abs/2307.15818}, 
}

@misc{kim2024openvla,
      title={OpenVLA: An Open-Source Vision-Language-Action Model}, 
      author={Moo Jin Kim and Karl Pertsch and Siddharth Karamcheti and Ted Xiao and Ashwin Balakrishna and Suraj Nair and Rafael Rafailov and Ethan Foster and Grace Lam and Pannag Sanketi and Quan Vuong and Thomas Kollar and Benjamin Burchfiel and Russ Tedrake and Dorsa Sadigh and Sergey Levine and Percy Liang and Chelsea Finn},
      year={2024},
      eprint={2406.09246},
      archivePrefix={arXiv},
      primaryClass={cs.RO},
      url={https://arxiv.org/abs/2406.09246}, 
}

@misc{physicalintelligence2025pi05,
      title={$\pi_{0.5}$: a Vision-Language-Action Model with Open-World Generalization}, 
      author={Physical Intelligence and Kevin Black and Noah Brown and James Darpinian and Karan Dhabalia and Danny Driess and Adnan Esmail and Michael Equi and Chelsea Finn and Niccolo Fusai and Manuel Y. Galliker and Dibya Ghosh and Lachy Groom and Karol Hausman and Brian Ichter and Szymon Jakubczak and Tim Jones and Liyiming Ke and Devin LeBlanc and Sergey Levine and Adrian Li-Bell and Mohith Mothukuri and Suraj Nair and Karl Pertsch and Allen Z. Ren and Lucy Xiaoyang Shi and Laura Smith and Jost Tobias Springenberg and Kyle Stachowicz and James Tanner and Quan Vuong and Homer Walke and Anna Walling and Haohuan Wang and Lili Yu and Ury Zhilinsky},
      year={2025},
      eprint={2504.16054},
      archivePrefix={arXiv},
      primaryClass={cs.LG},
      url={https://arxiv.org/abs/2504.16054}, 
}

@misc{hancock2025actionslanguage,
      title={Actions as Language: Fine-Tuning VLMs into VLAs Without Catastrophic Forgetting}, 
      author={Asher J. Hancock and Xindi Wu and Lihan Zha and Olga Russakovsky and Anirudha Majumdar},
      year={2025},
      eprint={2509.22195},
      archivePrefix={arXiv},
      primaryClass={cs.RO},
      url={https://arxiv.org/abs/2509.22195}, 
}

@misc{driess2025knowledgeinsulation,
      title={Knowledge Insulating Vision-Language-Action Models: Train Fast, Run Fast, Generalize Better}, 
      author={Danny Driess and Jost Tobias Springenberg and Brian Ichter and Lili Yu and Adrian Li-Bell and Karl Pertsch and Allen Z. Ren and Homer Walke and Quan Vuong and Lucy Xiaoyang Shi and Sergey Levine},
      year={2025},
      eprint={2505.23705},
      archivePrefix={arXiv},
      primaryClass={cs.LG},
      url={https://arxiv.org/abs/2505.23705}, 
}

@misc{zhao2025cotvla,
      title={CoT-VLA: Visual Chain-of-Thought Reasoning for Vision-Language-Action Models}, 
      author={Qingqing Zhao and Yao Lu and Moo Jin Kim and Zipeng Fu and Zhuoyang Zhang and Yecheng Wu and Zhaoshuo Li and Qianli Ma and Song Han and Chelsea Finn and Ankur Handa and Ming-Yu Liu and Donglai Xiang and Gordon Wetzstein and Tsung-Yi Lin},
      year={2025},
      eprint={2503.22020},
      archivePrefix={arXiv},
      primaryClass={cs.CV},
      url={https://arxiv.org/abs/2503.22020}, 
}

@misc{lee2025molmoact,
      title={MolmoAct: Action Reasoning Models that can Reason in Space}, 
      author={Jason Lee and Jiafei Duan and Haoquan Fang and Yuquan Deng and Shuo Liu and Boyang Li and Bohan Fang and Jieyu Zhang and Yi Ru Wang and Sangho Lee and Winson Han and Wilbert Pumacay and Angelica Wu and Rose Hendrix and Karen Farley and Eli VanderBilt and Ali Farhadi and Dieter Fox and Ranjay Krishna},
      year={2025},
      eprint={2508.07917},
      archivePrefix={arXiv},
      primaryClass={cs.RO},
      url={https://arxiv.org/abs/2508.07917}, 
}

@misc{liang2024mixture,
      title={Mixture-of-Transformers: A Sparse and Scalable Architecture for Multi-Modal Foundation Models}, 
      author={Weixin Liang and Lili Yu and Liang Luo and Srinivasan Iyer and Ning Dong and Chunting Zhou and Gargi Ghosh and Mike Lewis and Wen-tau Yih and Luke Zettlemoyer and Xi Victoria Lin},
      year={2025},
      eprint={2411.04996},
      archivePrefix={arXiv},
      primaryClass={cs.CL},
      url={https://arxiv.org/abs/2411.04996}, 
}

@misc{beyer2024paligemma,
      title={PaliGemma: A versatile 3B VLM for transfer}, 
      author={Lucas Beyer and Andreas Steiner and André Susano Pinto and Alexander Kolesnikov and Xiao Wang and Daniel Salz and Maxim Neumann and Ibrahim Alabdulmohsin and Michael Tschannen and Emanuele Bugliarello and Thomas Unterthiner and Daniel Keysers and Skanda Koppula and Fangyu Liu and Adam Grycner and Alexey Gritsenko and Neil Houlsby and Manoj Kumar and Keran Rong and Julian Eisenschlos and Rishabh Kabra and Matthias Bauer and Matko Bošnjak and Xi Chen and Matthias Minderer and Paul Voigtlaender and Ioana Bica and Ivana Balazevic and Joan Puigcerver and Pinelopi Papalampidi and Olivier Henaff and Xi Xiong and Radu Soricut and Jeremiah Harmsen and Xiaohua Zhai},
      year={2024},
      eprint={2407.07726},
      archivePrefix={arXiv},
      primaryClass={cs.CV},
      url={https://arxiv.org/abs/2407.07726}, 
}

@misc{zhou2019continuity,
      title={On the Continuity of Rotation Representations in Neural Networks}, 
      author={Yi Zhou and Connelly Barnes and Jingwan Lu and Jimei Yang and Hao Li},
      year={2020},
      eprint={1812.07035},
      archivePrefix={arXiv},
      primaryClass={cs.LG},
      url={https://arxiv.org/abs/1812.07035}, 
}

@misc{abc2026,
      title={Scalable Behavior Cloning with Open Data, Training, and Evaluation}, 
      author={Arthur Allshire and Himanshu Gaurav Singh and Ritvik Singh and Adam Rashid and Hongsuk Choi and David McAllister and Justin Yu and Yiyuan Chen and Huang Huang and Pieter Abbeel and Xi Chen and Rocky Duan and Phillip Isola and Jitendra Malik and Fred Shentu and Guanya Shi and Philipp Wu and Angjoo Kanazawa},
      year={2026},
      eprint={2606.27375},
      archivePrefix={arXiv},
      primaryClass={cs.RO},
      url={https://arxiv.org/abs/2606.27375}, 
}

@misc{lin2026systematiccotraining,
      title={A Systematic Study of Data Modalities and Strategies for Co-training Large Behavior Models for Robot Manipulation}, 
      author={Fanqi Lin and Kushal Arora and Jean Mercat and Haruki Nishimura and Paarth Shah and Chen Xu and Mengchao Zhang and Mark Zolotas and Maya Angeles and Owen Pfannenstiehl and Andrew Beaulieu and Jose Barreiros},
      year={2026},
      eprint={2602.01067},
      archivePrefix={arXiv},
      primaryClass={cs.RO},
      url={https://arxiv.org/abs/2602.01067}, 
}

@misc{mu2023embodiedgpt,
      title={EmbodiedGPT: Vision-Language Pre-Training via Embodied Chain of Thought}, 
      author={Yao Mu and Qinglong Zhang and Mengkang Hu and Wenhai Wang and Mingyu Ding and Jun Jin and Bin Wang and Jifeng Dai and Yu Qiao and Ping Luo},
      year={2023},
      eprint={2305.15021},
      archivePrefix={arXiv},
      primaryClass={cs.RO},
      url={https://arxiv.org/abs/2305.15021}, 
}

@misc{clark2025rad,
      title={Action-Free Reasoning for Policy Generalization}, 
      author={Jaden Clark and Suvir Mirchandani and Dorsa Sadigh and Suneel Belkhale},
      year={2025},
      eprint={2502.03729},
      archivePrefix={arXiv},
      primaryClass={cs.RO},
      url={https://arxiv.org/abs/2502.03729}, 
}

@misc{fu2025metis,
      title={METIS: Multi-Source Egocentric Training for Integrated Dexterous Vision-Language-Action Model}, 
      author={Yankai Fu and Ning Chen and Junkai Zhao and Shaozhe Shan and Guocai Yao and Pengwei Wang and Zhongyuan Wang and Shanghang Zhang},
      year={2025},
      eprint={2511.17366},
      archivePrefix={arXiv},
      primaryClass={cs.RO},
      url={https://arxiv.org/abs/2511.17366}, 
}

@misc{kim2026demian,
      title={How to Instruct Your Robot: Dense Language Annotations Power Robot Policy Learning}, 
      author={Bosung Kim and Ruiyi Wang and David Acuna and Jaehun Jung and Alexander Trevithick and Brandon Cui and Yejin Choi and Prithviraj Ammanabrolu},
      year={2026},
      eprint={2605.17077},
      archivePrefix={arXiv},
      primaryClass={cs.RO},
      url={https://arxiv.org/abs/2605.17077}, 
}

@misc{kareer2024egomimic,
      title={EgoMimic: Scaling Imitation Learning via Egocentric Video}, 
      author={Simar Kareer and Dhruv Patel and Ryan Punamiya and Pranay Mathur and Shuo Cheng and Chen Wang and Judy Hoffman and Danfei Xu},
      year={2024},
      eprint={2410.24221},
      archivePrefix={arXiv},
      primaryClass={cs.RO},
      url={https://arxiv.org/abs/2410.24221}, 
}

@misc{punamiya2025egobridge,
      title={EgoBridge: Domain Adaptation for Generalizable Imitation from Egocentric Human Data}, 
      author={Ryan Punamiya and Dhruv Patel and Patcharapong Aphiwetsa and Pranav Kuppili and Lawrence Y. Zhu and Simar Kareer and Judy Hoffman and Danfei Xu},
      year={2025},
      eprint={2509.19626},
      archivePrefix={arXiv},
      primaryClass={cs.RO},
      url={https://arxiv.org/abs/2509.19626}, 
}

@misc{zheng2026egoscale,
      title={EgoScale: Scaling Dexterous Manipulation with Diverse Egocentric Human Data}, 
      author={Ruijie Zheng and Dantong Niu and Yuqi Xie and Jing Wang and Mengda Xu and Yunfan Jiang and Fernando Castañeda and Fengyuan Hu and You Liang Tan and Letian Fu and Trevor Darrell and Furong Huang and Yuke Zhu and Danfei Xu and Linxi Fan},
      year={2026},
      eprint={2602.16710},
      archivePrefix={arXiv},
      primaryClass={cs.RO},
      url={https://arxiv.org/abs/2602.16710}, 
}

@misc{kim2026egopi,
      title={Ego-Pi: VLA Fine-Tuning for Ego-Centric Human and Robot Data}, 
      author={Ji Woong Kim and Ke Wang and Zipeng Fu and Sirui Chen and Cong Zhao and Jeff Lai and Chelsea Finn},
      year={2026},
      eprint={2606.08107},
      archivePrefix={arXiv},
      primaryClass={cs.RO},
      url={https://arxiv.org/abs/2606.08107}, 
}

@misc{liu2026oatpolicy,
      title={Ordered Action Tokens for Visuomotor Policy Learning}, 
      author={Chaoqi Liu and Yue Zhao and Haonan Chen and Xiaoshen Han and Jiawei Gao and Ehsan Adeli and Yilun Du},
      year={2026},
      eprint={2607.21670},
      archivePrefix={arXiv},
      primaryClass={cs.RO},
      url={https://arxiv.org/abs/2607.21670}, 
}

@misc{dong2026actioncodec,
      title={ActionCodec: What Makes for Good Action Tokenizers}, 
      author={Zibin Dong and Yicheng Liu and Shiduo Zhang and Baijun Ye and Yifu Yuan and Fei Ni and Jingjing Gong and Xipeng Qiu and Hang Zhao and Yinchuan Li and Jianye Hao},
      year={2026},
      eprint={2602.15397},
      archivePrefix={arXiv},
      primaryClass={cs.RO},
      url={https://arxiv.org/abs/2602.15397}, 
}

@misc{zawalski2024ecot,
      title={Robotic Control via Embodied Chain-of-Thought Reasoning}, 
      author={Michał Zawalski and William Chen and Karl Pertsch and Oier Mees and Chelsea Finn and Sergey Levine},
      year={2025},
      eprint={2407.08693},
      archivePrefix={arXiv},
      primaryClass={cs.RO},
      url={https://arxiv.org/abs/2407.08693}, 
}

@misc{chen2025ecotlite,
      title={Training Strategies for Efficient Embodied Reasoning}, 
      author={William Chen and Suneel Belkhale and Suvir Mirchandani and Oier Mees and Danny Driess and Karl Pertsch and Sergey Levine},
      year={2025},
      eprint={2505.08243},
      archivePrefix={arXiv},
      primaryClass={cs.RO},
      url={https://arxiv.org/abs/2505.08243}, 
}

@misc{sun2026ervla,
      title={Revisiting Embodied Chain-of-Thought for Generalizable Robot Manipulation}, 
      author={Nan Sun and Yuan Zhang and Yongkun Yang and Wentao Zhao and Peiyan Li and Jun Guo and Wenxuan Song and Pengxiang Ding and Runze Suo and Yifei Su and Xin Xiao and Xinghang Li and Huaping Liu},
      year={2026},
      eprint={2606.03784},
      archivePrefix={arXiv},
      primaryClass={cs.RO},
      url={https://arxiv.org/abs/2606.03784}, 
}

@misc{sun2025emmax,
      title={Emma-X: An Embodied Multimodal Action Model with Grounded Chain of Thought and Look-ahead Spatial Reasoning}, 
      author={Qi Sun and Pengfei Hong and Tej Deep Pala and Vernon Toh and U-Xuan Tan and Deepanway Ghosal and Soujanya Poria},
      year={2024},
      eprint={2412.11974},
      archivePrefix={arXiv},
      primaryClass={cs.RO},
      url={https://arxiv.org/abs/2412.11974}, 
}

@misc{yuan2026embodiedr1,
      title={Embodied-R1: Reinforced Embodied Reasoning for General Robotic Manipulation}, 
      author={Yifu Yuan and Haiqin Cui and Yaoting Huang and Yibin Chen and Fei Ni and Zibin Dong and Pengyi Li and Yan Zheng and Hongyao Tang and Jianye Hao},
      year={2026},
      eprint={2508.13998},
      archivePrefix={arXiv},
      primaryClass={cs.RO},
      url={https://arxiv.org/abs/2508.13998}, 
}

@misc{cai2025innonscalingegocentricmanipulation,
      title={In-N-On: Scaling Egocentric Manipulation with in-the-wild and on-task Data}, 
      author={Xiongyi Cai and Ri-Zhao Qiu and Geng Chen and Lai Wei and Isabella Liu and Tianshu Huang and Xuxin Cheng and Xiaolong Wang},
      year={2025},
      eprint={2511.15704},
      archivePrefix={arXiv},
      primaryClass={cs.RO},
      url={https://arxiv.org/abs/2511.15704}, 
}

@misc{hoque2026egodexlearningdexterousmanipulation,
      title={EgoDex: Learning Dexterous Manipulation from Large-Scale Egocentric Video}, 
      author={Ryan Hoque and Peide Huang and David J. Yoon and Mouli Sivapurapu and Jian Zhang},
      year={2026},
      eprint={2505.11709},
      archivePrefix={arXiv},
      primaryClass={cs.CV},
      url={https://arxiv.org/abs/2505.11709}, 
}

@misc{grover2025enhancinggeneralizationvisionlanguageactionmodels,
      title={Enhancing Generalization in Vision-Language-Action Models by Preserving Pretrained Representations}, 
      author={Shresth Grover and Akshay Gopalkrishnan and Bo Ai and Henrik I. Christensen and Hao Su and Xuanlin Li},
      year={2025},
      eprint={2509.11417},
      archivePrefix={arXiv},
      primaryClass={cs.RO},
      url={https://arxiv.org/abs/2509.11417}, 
}

@misc{kareer2025emergencehumanrobottransfer,
      title={Emergence of Human to Robot Transfer in Vision-Language-Action Models}, 
      author={Simar Kareer and Karl Pertsch and James Darpinian and Judy Hoffman and Danfei Xu and Sergey Levine and Chelsea Finn and Suraj Nair},
      year={2025},
      eprint={2512.22414},
      archivePrefix={arXiv},
      primaryClass={cs.RO},
      url={https://arxiv.org/abs/2512.22414}, 
}

@misc{wei2026psi0openfoundationmodel,
      title={$\Psi_0$: An Open Foundation Model Towards Universal Humanoid Loco-Manipulation}, 
      author={Songlin Wei and Hongyi Jing and Boqian Li and Zhenyu Zhao and Jiageng Mao and Zhenhao Ni and Sicheng He and Jie Liu and Xiawei Liu and Kaidi Kang and Sheng Zang and Weiduo Yuan and Marco Pavone and Di Huang and Yue Wang},
      year={2026},
      eprint={2603.12263},
      archivePrefix={arXiv},
      primaryClass={cs.RO},
      url={https://arxiv.org/abs/2603.12263}, 
}

@misc{shao2024deepseekmath,
      title={DeepSeekMath: Pushing the Limits of Mathematical Reasoning in Open Language Models},
      author={Zhihong Shao and Peiyi Wang and Qihao Zhu and Runxin Xu and Junxiao Song and Xiao Bi and Haowei Zhang and Mingchuan Zhang and Y. K. Li and Y. Wu and Daya Guo},
      year={2024},
      eprint={2402.03300},
      archivePrefix={arXiv},
      primaryClass={cs.CL},
      url={https://arxiv.org/abs/2402.03300},
}

@misc{schulman2017ppo,
      title={Proximal Policy Optimization Algorithms},
      author={John Schulman and Filip Wolski and Prafulla Dhariwal and Alec Radford and Oleg Klimov},
      year={2017},
      eprint={1707.06347},
      archivePrefix={arXiv},
      primaryClass={cs.LG},
      url={https://arxiv.org/abs/1707.06347},
}

@misc{ye2026worldactionmodelszeroshot,
      title={World Action Models are Zero-shot Policies}, 
      author={Seonghyeon Ye and Yunhao Ge and Kaiyuan Zheng and Shenyuan Gao and Sihyun Yu and George Kurian and Suneel Indupuru and You Liang Tan and Chuning Zhu and Jiannan Xiang and Ayaan Malik and Kyungmin Lee and William Liang and Nadun Ranawaka and Jiasheng Gu and Yinzhen Xu and Guanzhi Wang and Fengyuan Hu and Avnish Narayan and Johan Bjorck and Jing Wang and Gwanghyun Kim and Dantong Niu and Ruijie Zheng and Yuqi Xie and Jimmy Wu and Qi Wang and Ryan Julian and Danfei Xu and Yilun Du and Yevgen Chebotar and Scott Reed and Jan Kautz and Yuke Zhu and Linxi "Jim" Fan and Joel Jang},
      year={2026},
      eprint={2602.15922},
      archivePrefix={arXiv},
      primaryClass={cs.RO},
      url={https://arxiv.org/abs/2602.15922}, 
}
\bibliographystyle{unsrtnat}

\clearpage
\beginappendix{
\appendix

\section{Training Details}
\label{app:training_details}

\subsection{Model Architecture and Tensor Shapes}
\label{app:model_architecture}

\begin{figure}[H]
  \centering
  \includegraphics[width=0.99\textwidth]{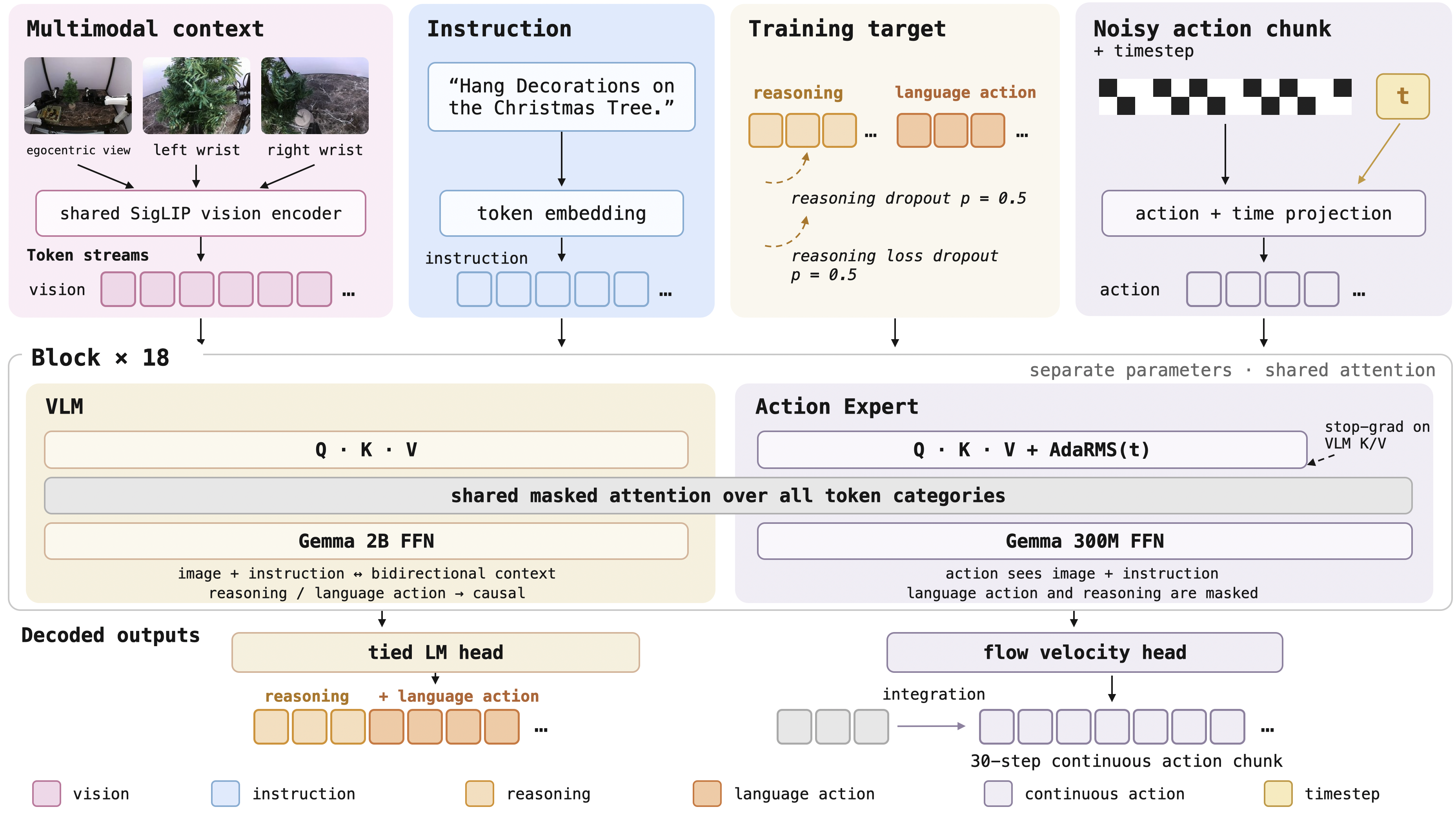}
  \caption{\textbf{Detailed EgoLAP architecture.} A shared SigLIP encoder maps each camera view to visual tokens. The 2B VLM branch predicts motion-level reasoning and the language action; the 300M action expert maps a noisy action chunk to its flow velocity. The two branches hold separate parameters but join the same masked-attention operation at every transformer layer. Action tokens attend to the observation, instruction, and discretized state, never to the textual targets, and gradients through the VLM keys and values are stopped.}
  \label{fig:detailed_model_arch}
\end{figure}

Let $B$ be the batch size and $V$ the number of enabled camera slots. Before tokenization, a sample contains RGB observations
$o_t\in\mathbb{R}^{B\times V\times224\times224\times3}$, a task instruction, a normalized proprioceptive state $s_t\in\mathbb{R}^{B\times20}$, and a continuous target $a_{t:t+30}\in\mathbb{R}^{B\times30\times14}$. The 20 state values comprise ten values per arm: end-effector position, a continuous 6D rotation, and gripper state. The 14 action values comprise seven Cartesian-control values per arm; single-arm samples occupy one seven-dimensional block and pad the other. For samples with fewer than three views, absent views are zero-filled and masked.

Each $224\times224$ image produces 256 SigLIP patch tokens. The normalized state is discretized into 256 uniform bins and serialized with the instruction, so it is part of the text prefix rather than a separate action-expert token. For EgoLAP, the padded text tensor has shape $B\times320$. Thus, with three views, the maximum VLM prefix has shape $B\times(3\cdot256+320)\times2048$. Padding and absent-image tokens are removed by the attention mask. The VLM has 18 transformer blocks, width 2048, an FFN width of 16,384, eight query heads, and one key--value head.

The action expert receives $a_{t:t+30}^{\tau}\in\mathbb{R}^{B\times30\times14}$ and projects it to $B\times30\times1024$. A sinusoidal embedding of $\tau$ produces one $B\times1024$ conditioning vector used by adaptive RMS normalization in all 18 expert blocks. The expert has FFN width 4096, eight query heads, and one key--value head. Its velocity head returns $v_\phi\in\mathbb{R}^{B\times30\times14}$, exactly matching the target action-chunk shape. The tied VLM head produces token logits of shape $B\times L\times257{,}152$; only reasoning and action-target positions receive text loss. At inference, ten Euler steps integrate a noise tensor of shape $B\times30\times14$ to the final action chunk of the same shape. Text is not decoded for control.

\subsection{Prompt and Language-Action Format}
\label{app:prompt_action_format}

\noindent\textbf{Prompt serialization.}
The observation images are followed by a single text sequence with the literal format
\begin{quote}
\small\ttfamily
Task: [instruction], predict the robot's action in the egocentric frame; State: [$b_1$ $\ldots$ $b_d$]; Answer:
\end{quote}
The instruction is stripped of a trailing period, and underscores and line breaks are replaced by spaces. Each $b_i\in\{0,\ldots,255\}$ is a uniformly discretized normalized state value. Trailing zero-padded state dimensions are omitted, leaving $d{=}10$ values for a single arm and $d{=}20$ for bimanual robot and human samples. 
The reference run reuses the exacting phrase \texttt{egocentric frame} and \texttt{robot's action} for all embodiments, including human samples.

For a sample without motion-level reasoning, the language action $m_t$ immediately follows \texttt{Answer:}. When a rationale is included, the complete target is
\begin{quote}
\small\ttfamily
Answer: [$r_t$] Concrete actions to take are: [$m_t$]
\end{quote}
The boundary phrase is fixed and separates the optional rationale from the language action, allowing their losses to be masked independently while retaining one causal target sequence. At inference, the policy receives the same observation--instruction--state prefix, but the continuous action expert does not decode either textual field.

\noindent\textbf{Language-action grammar.}
Each arm is represented by an ordered, comma-separated clause. Translation is emitted in the order longitudinal, vertical, then lateral; rotation follows as roll, pitch, then yaw; and the clause always ends with the final gripper state. Table~\ref{tab:language_action_vocabulary} gives the exact vocabulary.

\begin{table}[H]
\centering
\footnotesize
\caption{Canonical component vocabulary for language actions. Here $n$ is an integer number of centimeters and $\theta$ is an integer multiple of $5$ degrees. A component is omitted when its rounded magnitude is zero.}
\label{tab:language_action_vocabulary}
\renewcommand{\arraystretch}{1.12}
\begin{tabular}{@{}l l l@{}}
\toprule
\textbf{Component} & \textbf{Positive direction} & \textbf{Negative direction} \\
\midrule
Translation $x$ & \texttt{move forward $n$ cm} & \texttt{move back $n$ cm} \\
Translation $z$ & \texttt{move up $n$ cm} & \texttt{move down $n$ cm} \\
Translation $y$ & \texttt{move left $n$ cm} & \texttt{move right $n$ cm} \\
Roll & \texttt{tilt left $\theta$ degrees} & \texttt{tilt right $\theta$ degrees} \\
Pitch & \texttt{tilt back $\theta$ degrees} & \texttt{tilt forward $\theta$ degrees} \\
Yaw & \texttt{rotate counterclockwise $\theta$ degrees} & \texttt{rotate clockwise $\theta$ degrees} \\
Gripper state & \texttt{open gripper} & \texttt{close gripper} \\
\bottomrule
\end{tabular}
\end{table}

Translational displacement is rounded to integer centimeters, and each Euler-angle component is rounded to the nearest $5$ degrees. The open/close token describes the final aperture state rather than a displacement and is therefore retained even when all pose components are omitted. The embodiment-specific serializations are as follows.

\noindent\emph{Single-arm robot.}
One clause is emitted, with its arm prefix sampled uniformly during training:
\begin{quote}
\small\ttfamily
[Left$\mid$Right] arm: [motion components], [open$\mid$close] gripper
\end{quote}
For example, a DROID sample in the reference run is serialized as \texttt{Left arm: move forward 8 cm, move down 6 cm, move right 7 cm, tilt forward 10 degrees, rotate counterclockwise 10 degrees, close gripper}.

\noindent\emph{Bimanual robot.}
Both clauses are emitted in a fixed order and separated by a period:
\begin{quote}
\small\ttfamily
Left arm: [left components]. Right arm: [right components]
\end{quote}

\noindent\emph{Egocentric human.}
Human hand trajectories use the same bimanual template rather than a separate human-only vocabulary. \texttt{Left arm} and \texttt{Right arm} identify the tracked hands, while \texttt{open gripper} and \texttt{close gripper} encode the corresponding binary hand-aperture state.
This shared serialization provides a common interface through which human and robot motion supervise the same VLM tokens.

\subsection{Flow-Matching Objective}
\label{app:flow_matching}

For a continuous action chunk $a_{t:t+K}\in\mathbb{R}^{K\times D}$, Gaussian noise $\epsilon\sim\mathcal{N}(0,I)$, and interpolation time $\tau=0.001+0.999z$ with $z\sim\operatorname{Beta}(1.5,1)$, define the interpolated state and target velocity:
\begin{equation}
    a_{t:t+K}^\tau=(1-\tau)a_{t:t+K}+\tau\epsilon,
    \qquad u_t=\epsilon-a_{t:t+K} .
\end{equation}
The action expert $v_\phi$ is trained to recover this velocity:
\begin{equation}
\mathcal{L}_{\mathrm{FM}}=
  \mathbb{E}_{\tau,\epsilon}
  \left[\frac{1}{KD}\left\|v_\phi\!\left(a_{t:t+K}^\tau,\tau,
  \operatorname{sg}[h_\theta(x_t)]\right)-u_t\right\|_F^2\right],
\label{eq:flow_matching}
\end{equation}
where $x_t=(o_t,\ell,s_t)$ is the multimodal prefix, $h_\theta$ is the backbone prefix representation and $\operatorname{sg}$ denotes stop-gradient, implementing the knowledge-insulation boundary~\citep{driess2025knowledgeinsulation} that prevents continuous-action gradients from eroding VLM's pretrained representation. At inference, actions are generated by integrating the learned velocity field from $\tau{=}1$ (noise) to $\tau{=}0$ (action) using a fixed-step ODE solver.

\subsection{Stochastic supervision}
\label{app:stochastic_supervision}

To avoid VLMs overfitting to reasoning, we stochastically vary whether reasoning appears in the target and whether it receives loss. For rationale tokens $r_t=(r_{t,1},\ldots,r_{t,M})$ and discrete action tokens $m_t=(m_{t,1},\ldots,m_{t,N})$, draw two independent variables
\begin{equation}
    B\sim\operatorname{Bernoulli}(p_{\mathrm{ctx}}),
    \qquad
    C\sim\operatorname{Bernoulli}(p_{\mathrm{sup}}).
    \label{eq:reasoning_masks}
\end{equation}
If $B{=}1$, the rationale is inserted as a teacher-forced causal prefix before $m_t$; otherwise, the VLM predicts $m_t$ directly. Write $r_t^{(B)}=r_t$ when $B{=}1$ and $r_t^{(B)}=\varnothing$ otherwise. Conditional on the rationale's presence, $C$ determines whether its tokens receive language-modeling loss. The masked per-sample objective is
\begin{align}
\ell_{\mathrm{text}}(B,C)
=-\frac{1}{N+BCM}\Bigg[
&BC\sum_{j=1}^{M}\log p_\theta\!\left(
  r_{t,j}\mid x_t,r_{t,<j}\right) \notag\\[-1pt]
&+\sum_{i=1}^{N}\log p_\theta\!\left(
  m_{t,i}\mid x_t,r_t^{(B)},m_{t,<i}\right)
\Bigg].
\label{eq:text_objective}
\end{align}
The text term in Eq.~\ref{eq:full_objective} is $\mathcal{L}_{\mathrm{text}}=\mathbb{E}_{B,C}[\ell_{\mathrm{text}}(B,C)]$. We use $p_{\mathrm{ctx}}=p_{\mathrm{sup}}=0.5$; for samples without a rationale, we set $B=C=0$. Thus, among labeled samples, half train direct language action prediction, one quarter use reasoning only as causal context, and one quarter both condition on and supervise the rationale. The discrete action tokens are supervised in every case.

\subsection{Motion-Level Reasoning Annotation}
\label{app:reasoning_annotation}

\noindent\textbf{Coverage.}
Motion-level reasoning labels exist for only two of the datasets in our mixture: the MECKA subset of EgoVerse~\citep{punamiya2026egoverseegocentrichumandataset}, which supplies all of our egocentric human reasoning, and DROID~\citep{khazatsky2024droid}, which supplies all of our robot reasoning. Together they account for $32\%$ of each training batch (Fig.~\ref{fig:data_mixture}). Every other dataset --- ABC YAM, the remaining bimanual robot datasets, the Scale and Aria subsets of EgoVerse, and the remaining OXE datasets --- contributes language actions and continuous actions but no rationale. Those samples take $B{=}C{=}0$ in Eq.~\ref{eq:text_objective} and are therefore trained exactly as in the no-reasoning variants. The same two annotated sources are used for every reasoning policy we report, including EgoFAST and EgoOAT, so reasoning coverage is never a confound between action representations.

\noindent\textbf{Annotation procedure.}
We generate the annotations offline with \texttt{gemini-3.5-flash} for DROID and \texttt{gemini-3.7-flash} for MECKA. Both runs use temperature 0. Gemini is given the task instruction and a temporally ordered set of action anchors. The target language action is an input to the annotation procedure to provide Gemini with the action context, which we empirically find to lead to more grounded reasoning labels compared to only providing images. Appendix~\ref{app:reasoning_examples} shows representative DROID and MECKA annotations together with the labeling prompt.

For DROID, anchors are formed from adaptive action segments with a nominal stride of 15 filtered steps (approximately one second). A segment can end early at a persistent gripper-state change or a large change in translational direction. At each anchor, Gemini receives the current primary and wrist-camera images, the current gripper and transition metadata, the language action computed from the numeric actions over that segment, and a compact text-only summary of up to five seconds of preceding actions. The prompt distinguishes the state visible at the anchor from the command to be executed over the following segment, which prevents a future grasp, release, or contact event from being described as already complete.

For MECKA, subsequences belonging to the same high-level recording are first concatenated in their recorded order. Then, adaptive anchors are selected with a separation of 15--30 frames at 30\,Hz. Each anchor contains the head-mounted egocentric image, the high-level instruction and stitched subtask context, the segment-level bimanual language action, and the same compact action history. Since MECKA records human manipulation, the prompt maps the shared action vocabulary to left and right hands, fingers, contact, support, and release rather than robot grippers.

Gemini outputs a trajectory summary, a subtask plan, and, for every requested anchor, the completed, current, and next subtasks; observation--action grounding; the main rationale; a visual-relation sentence; a motion-effect sentence; and an alternative-motion diagnostic. However, during training, we only use the main motion-level reasoning rationale.

\subsection{Training Hyperparameters}
\label{app:hyperparameters}

Table~\ref{tab:hyperparams} summarizes the pre-training hyperparameters. We train all policy variants for 40k steps with the same hyperparameter settings. This includes EgoLAP, EgoFAST, and EgoOAT, each with and without reasoning, as well as the robot-data-only ablation. The language action loss weight is set below 1.0 because the discrete token loss objective converges faster than the flow-matching expert; slightly downweighting this loss prevents the backbone from overfitting early in training. Appendix~\ref{app:baseline} reports additional observations under varying training budgets.

\begin{table}[H]
\centering
\footnotesize
\caption{Shared policy pre-training hyperparameters.}
\label{tab:hyperparams}
\renewcommand{\arraystretch}{1.15}
\begin{tabular}{l c}
\toprule
\textbf{Hyperparameter} & \textbf{Value} \\
\midrule
Batch size & 2048 \\
Hardware & 64 $\times$ TPU v6e \\
Training steps & 40k \\
Learning-rate schedule & 2500-step linear warmup, then constant \\
Peak learning rate & $5 \times 10^{-5}$ \\
Optimizer & Adam ($\beta_1{=}0.9$, $\beta_2{=}0.95$, $\epsilon{=}10^{-8}$) \\
Weight decay & $1 \times 10^{-4}$ \\
Gradient clipping norm & 1.0 \\
EMA decay & 0.999 \\
Language-action loss weight & 0.8 \\
Flow-matching loss weight & 1.0 \\
Action chunk length & 30 steps \\
Action dimension & 14 \\
State dimension & 20 \\
Image resolution & $224 \times 224$ \\
Numeric precision & bfloat16 \\
Action/state normalization & per-category 1st--99th percentile bounds \\
Flow-time distribution & $0.001+0.999\operatorname{Beta}(1.5,1)$ \\
Inference integration steps & 10 \\
Wrist-image dropout & 0.1 \\
Random camera masking & 0.2 \\
Validation/checkpoint interval & 1000 steps \\
Random seed & 0 \\
\bottomrule
\end{tabular}
\end{table}

\noindent \textbf{Image pre-processing and augmentation.} Input images are resized with aspect-ratio-preserving symmetric padding to a fixed resolution (224$\times$224 in the default configuration). During training, image augmentation is enabled by default for each model-facing camera view. Images are mapped from $[-1,1]$ to $[0,1]$ and transformed using a random crop of $\lfloor0.95W\rfloor\times\lfloor0.95H\rfloor$, resizing back to the target resolution, a random in-plane rotation sampled uniformly from $[-5^\circ,5^\circ]$, and Augmax color jitter configured with brightness, contrast, and saturation strengths of 0.2. The augmented images are then mapped back to $[-1,1]$. Random seeds are independent across batch elements but shared across camera views for a given element.

\subsection{Data Mixture}
\label{app:data_mixture}

Figure~\ref{fig:data_mixture} shows the training data mixture reflecting the composition of each training batch. The mixture spans four groups: ABC robot data~\citep{abc2026} (30\%, from the ABC YAM platform), egocentric human data from EgoVerse~\citep{punamiya2026egoverseegocentrichumandataset} (25\%, drawn from the MECKA, Scale, and Aria subsets), additional bimanual robot data (25\%, including AgiBot, Galaxea R1~Lite, and MolmoAct2 YAM), and single-arm OXE robot datasets~\citep{openx2023} (20\%, predominantly DROID~\citep{khazatsky2024droid} at 12\%). Of these, only MECKA and DROID carry motion-level reasoning labels, as described in Appendix~\ref{app:reasoning_annotation}. The total raw data amounts to around 10k hours.

\begin{figure}[H]
  \centering
  \includegraphics[width=\columnwidth]{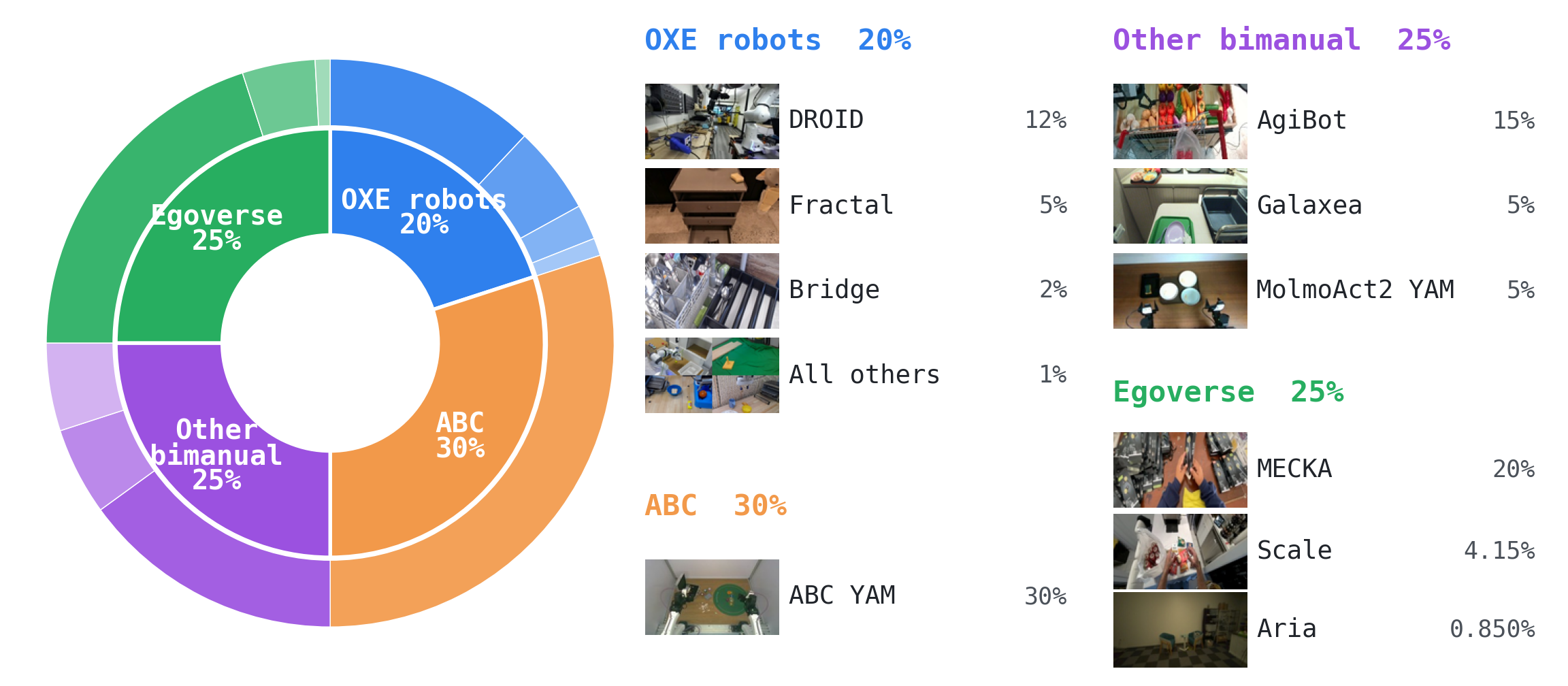}
  \caption{\textbf{Training data mixture.} Inner ring: group weights; outer ring: individual dataset contributions within each group. Only MECKA ($20\%$) and DROID ($12\%$) carry motion-level reasoning labels; the remaining samples are supervised with language actions and continuous actions alone.}
  \label{fig:data_mixture}
\end{figure}

\subsection{Baseline Implementation Details}
\label{app:baseline}

All matched policies use the architecture in Fig.~\ref{fig:detailed_model_arch}, the continuous $30\times14$ flow-matching target, the data mixture in Fig.~\ref{fig:data_mixture}, and the optimization settings in Table~\ref{tab:hyperparams}. FAST and OAT change only the autoregressive action target. Their action tokens supervise the VLM branch; they do not replace the continuous action expert. The expert therefore receives the same observations and continuous targets for every representation.

\noindent\textbf{FAST.}
We use the released FAST tokenizer~\citep{pertsch2025fast} to encode each normalized $30\times14$ action chunk. In the no-reasoning policy, the FAST tokens directly follow the observation--instruction--state prefix. In the reasoning policy, the motion rationale is placed before the codes. The tokenizer is fixed while the policy is trained.

\noindent\textbf{OAT.}
We separately train the OAT codec~\citep{liu2026oatpolicy} on normalized \(30\times14\) action chunks sampled from the same mixture used for policy pretraining, and freeze it during policy training. In our configuration, the codec represents each action chunk with a maximum-length sequence of 16 ordered discrete tokens. Since OAT is trained with power-of-two prefix dropout, any prefix of \(1\), \(2\), \(4\), \(8\), or \(16\) tokens can reconstruct the complete action chunk, with later tokens progressively refining the reconstruction. The encoder and decoder have hidden width 256, eight attention heads, and six Transformer blocks each, with dropout 0.1. Each latent is quantized using finite scalar quantization with levels \((8,8,6,5)\), yielding \(8\times8\times6\times5=1{,}920\) possible codes per token. We train the codec for 100,000 updates and use the same frozen checkpoint for both OAT policy variants. In the reasoning variant, reasoning tokens precede the OAT action tokens; the no-reasoning variant predicts the OAT action tokens directly.

\noindent\textbf{ECoT-style reasoning}~\citep{zawalski2024ecot}.
We implement the composite baseline in Sec.~\ref{subsec:reasoning_formats} as an ordered prefix containing subtask context, visible-object boxes, and a visual trace, followed by the same language action used by EgoLAP. The subtask context is reused from the offline Gemini annotations of Appendix~\ref{app:reasoning_annotation}: for DROID, we extract the standardized full-plan and current-subtask sentences from each annotated segment; for MECKA, we read the corresponding structured fields after stitching subsequences from the same recording. We obtain object boxes in a separate offline grounding pass: given an anchor image, the task instruction, and the current subtask, \texttt{gemini-3.5-flash} at temperature 0 returns task-relevant visible objects and their normalized boxes. We ground every second annotated anchor and do not propagate boxes to intervening frames. We map the boxes through the same aspect-preserving resize and symmetric padding as the input image, quantize them into the 1,024 PaliGemma location bins, and serialize each box in \((y_{\min},x_{\min},y_{\max},x_{\max})\) order.

The visual traces are computed from trajectories and camera parameters directly, rather than generated by Gemini. For DROID, we project the recorded end-effector positions into the calibrated exterior camera using its intrinsics and camera-to-base extrinsics; for MECKA, we project the left- and right-palm positions into the head-camera view using the synchronized head pose and camera intrinsics. In both cases, the target contains five future image-space positions sampled at 5\,Hz over the next second. Each \((y,x)\) position is quantized into the same 1,024 location-token bins. A box or trace is attached only when its source annotation or complete future projection is valid; otherwise that component is omitted without removing the training sample. The resulting target has the form \texttt{Full subtask plan is ... Current subtask is ... Visible objects: ... Visual motion plan ... Concrete actions to take are: ...}. It deliberately excludes the motion-level rationale, so the comparison isolates ECoT-style \emph{what}/\emph{where} supervision from our physical \emph{why} supervision. We retain loss on the ECoT-prefix tokens with probability 0.5 per sample while supervising the language-action tokens in every case. For the ECoT + motion-level reasoning baseline, we additionally appends motion-level reasoning after the visual traces.

\noindent\textbf{Reasoning and no-reasoning variants.}
The no-reasoning variants remove the rationale from the target and retain the action target on every sample. For reasoning variants, a rationale is included with probability 0.5. EgoLAP and EgoFAST retain rationale-token loss with probability 0.5 when the rationale is present. Language-action, FAST, or OAT action-target tokens receive loss in every case.

\noindent\textbf{Robot-only ablation.}
The robot-only baseline removes EgoVerse data while preserving the relative weights of every robot dataset in the full mixture; the remaining weights are renormalized. It is otherwise identical to the main model in architecture, hyperparameters, and compute budget.

\noindent\textbf{Checkpoint selection.}
All policies train for the same 40k-step budget, and we select one checkpoint per variant with a single validation procedure applied identically to every variant --- taking checkpoint with the lowest offline validation loss on the held out split.. The reported comparison point is therefore around 40k steps, except for EgoLAP, which we train for 30k steps only due to faster convergence. 

\subsection{Reinforcement Learning Experiment Details}
\label{app:rl}

This section specifies the language-action GRPO post-training reported in Sec.~\ref{subsec:finetuning_efficiency}. All three methods share the reward, the objective, the optimizer, the data pipeline, and the compute budget; they differ only in the SFT checkpoint they start from and in how rollouts are sampled.

\noindent\textbf{Datasets.}
Training draws from the ABC YAM dataset~\citep{abc2026}, from which we use the 1\% validation held out split from pre-training. Each eligible episode contributes exactly one observation per run.

\noindent\textbf{Reward.}
For arm $b\in\{\mathrm{left},\mathrm{right}\}$, let $p_b$ be the parsed translation in centimeters, $R_b$ the rotation reconstructed from the parsed XYZ Euler angles, and $g_b$ the binary gripper command, with $\star$ denoting the target. Writing the target-relative rotation vector as $\omega_b=\operatorname{Log}\!\big((R_b^{\star})^{-1}R_b\big)^{\vee}\in\mathbb{R}^3$ in radians, the cost and reward are
\begin{equation}
C=\sum_{b}\left[
\frac{\lVert p_b-p_b^{\star}\rVert_2^2}{(2\,\mathrm{cm})^2}
+\frac{\lVert \omega_b\rVert_2^2}{(\pi/18)^2}
+\mathbf{1}\!\left[g_b\neq g_b^{\star}\right]\right],
\qquad r=-C .
\label{eq:rl_reward}
\end{equation}
Terms are summed over arms, not averaged, so a $2$\,cm single-axis translation error, a $10^\circ$ relative orientation error, and one wrong gripper each cost one unit. Rotation error is the $SO(3)$ residual rather than a difference of Euler angles, and the gripper term is a discrete mismatch penalty rather than a cross-entropy. Valid rewards are non-positive and uncapped; an exact match scores zero.

A completion is valid only if the parser recovers exactly one clause for each arm with finite motion values; missing, duplicated, or non-finite clauses are rejected. Unspecified axes take the parser's zero default and an unspecified or ``maintain'' gripper takes its closed-state default. Within each group, an invalid completion receives $r_{\mathrm{invalid}}=-\max\!\big(1001,\,1+\max_{j\,\mathrm{valid}}C_j\big)$, which ranks it below every valid completion in the same group; a group in which all completions are invalid therefore yields zero reward-derived advantage. There is no reward term for rationale quality, response length, entropy, or simulator success.

\noindent\textbf{GRPO objective.}
Each update samples 32 observations and 64 completions per observation, and all 2{,}048 completions enter the gradient: the group is used for group-relative advantage estimation, not for selecting a single best response to imitate. Rewards and advantages are computed on the host without differentiating through the parser. For observation $i$ and completion $j$,
\begin{equation}
A_{ij}=\operatorname{clip}\!\left(\frac{r_{ij}-\bar r_i}{\max(\sigma_i,10^{-4})},\,-4,\,4\right),
\end{equation}
where $\bar r_i$ and $\sigma_i$ are the mean and population standard deviation within the group; groups whose reward range is at most $10^{-6}$ receive zero advantage. Let $P_\theta$ denote the policy's sequence distribution. For each retained response token, define the importance ratio $\rho_{ijt}$ against the sampling policy $P_{\mathrm{old}}$ and the reference log-ratio $\delta_{ijt}=\log P_{\mathrm{ref}}-\log P_\theta$. The minimized loss averages over tokens within a response and then over responses,
\begin{equation}
\mathcal{L}_{\mathrm{RL}}=\frac{1}{2048}\sum_{i,j}\frac{1}{L_{ij}}\sum_{t=1}^{L_{ij}}
\Big[-\min\!\big(\rho_{ijt}A_{ij},\ \operatorname{clip}(\rho_{ijt},1-\varepsilon,1+\varepsilon)A_{ij}\big)
+\beta\big(e^{\delta_{ijt}}-1-\delta_{ijt}\big)\Big],
\label{eq:rl_objective}
\end{equation}
with $\varepsilon=0.2$ for the clipped surrogate~\citep{schulman2017ppo} and a non-negative sampled (``$k_3$'') penalty of weight $\beta=0.02$ against the frozen SFT reference. The complete continuation --- any rationale, the language action, and the first EOS --- carries the same response-level advantage; prompt and post-EOS tokens are masked. Only VLM backbone (excluding the vision encoder) is updated; the vision encoder and the action expert are frozen. There is no auxiliary flow-matching, SFT, rationale, or entropy term.

\noindent\textbf{Optimization and compute.}
Table~\ref{tab:rl_hyperparams} lists the settings shared by all three runs.

\begin{table}[H]
\centering
\footnotesize
\caption{Shared language-action GRPO settings.}
\label{tab:rl_hyperparams}
\renewcommand{\arraystretch}{1.15}
\begin{tabular}{l c}
\toprule
\textbf{Setting} & \textbf{Value} \\
\midrule
Observations per update & 32 \\
Completions per observation & 64 (2{,}048 responses per update) \\
Gradient accumulation & 16 microbatches of 128 responses, one step per rollout bank \\
Optimizer & AdamW ($\beta_1{=}0.9$, $\beta_2{=}0.95$, $\epsilon{=}10^{-8}$), weight decay 0 \\
Learning rate & linear warmup $5\times10^{-8}\!\rightarrow\!10^{-6}$ over 20 updates, then constant \\
Gradient clipping & global norm 1.0, after accumulation \\
PPO clip range $\varepsilon$ & 0.2 \\
Reference penalty $\beta$ & 0.02, against the frozen SFT reference \\
Advantage & group-normalized, clipped to $[-4,4]$ \\
Rollout sampling & categorical, temperature 1, no top-$k$ or top-$p$ \\
Generation buffer & 211 response tokens (including any seeded prefix) \\
Retained response & through the first EOS, within the 320-token text buffer \\
Trainable parameters & language backbone and tied embeddings only \\
Hardware & eight TPU~v4 hosts (v4-64), 32 devices, FSDP size 32 \\
Precision & bfloat16 model; float32 log-softmax and optimizer moments \\
\bottomrule
\end{tabular}
\end{table}

\noindent\textbf{Validation and checkpoint selection.}
Every 20 updates, and at initialization and termination, we evaluate on a fixed development set of 224 observations drawn from 28 held-out ABC dataset episodes (eight stratified temporal bins per episode), disjoint from the RL training episodes. We log (i) a language-action metric --- eight temperature-1 samples plus one greedy decode per observation, scored with the same reward of Eq.~\ref{eq:rl_reward}, including invalid penalties --- and (ii) a continuous action-chunk metric from the flow-matching action expert: four action chunks per observation, scored as normalized MSE against the recorded command chunk. We manually choose the checkpoint that roughly minimizes both metrics.

\section{Evaluation Details}
\label{app:evaluation_details}

\noindent\textbf{Simulation} (Fig.~\ref{fig:results}~(a)).
The bimanual simulation benchmark contains 14 tasks grouped into three object-picking, eight object-placement, two color-selection, and one cloth-manipulation task categories. Picking covers colored blocks and fruit in isolated and cluttered scenes. Placement requires depositing fruit or blocks into bins or positioning them on plates and trays, including variants with distractors. Color tasks include selecting an object from an color-based instruction, and stacking one colored block on another. The cloth task requires folding a towel. Each task contains 100 episodes per checkpoint, yielding 1,400 episodes for a complete evaluation.

\noindent\textbf{Real robot}  (Fig.~\ref{fig:results}~(b)).
We evaluate on five real-world tasks on a custom bimanual YAM platform: two YAMs are mounted on the table, separated by around 0.8m.  We use realsense D435 as the base camera and realsense D405 for the two wrist cameras. Each task is decomposed into ordered stages, and a trial is scored by how many stages it completes: if task $i$ has $K_i$ stages, its normalized progress is $p_i=K_i^{-1}\sum_{k=1}^{K_i}s_{ik}$ with $s_{ik}\in\{0,1\}$. We report the mean of $p_i$ over trials. Object and container poses are randomized on the table between trials, and every policy sees the same initial configurations.

\begin{itemize}[leftmargin=1.5em,itemsep=3pt,topsep=3pt]
\item \textbf{Pick Place (Simple)}: place a toy banana into a bowl, with the banana and the bowl randomly positioned on the table. Three stages: pick up the banana, move it to the bowl, and release it into the bowl. 10 trials, split over two different objects.

\item \textbf{Pick Place (Complex)}: a cluttered tabletop holds several candidate objects and several candidate containers, and the instruction names one of each. The instructed pair may fall across the two arms' workspaces, in which case the task additionally requires a handover. Four stages: approach the instructed object, acquire and lift it, transport it to the instructed container, and place it. 12 trials, split over four different objects.

\item \textbf{Fold Pants}: fold a pair of pants in half. Five stages: grasp one end of the garment, grasp both ends, manipulate or lift the ends, produce a partial overlap, and complete the fold. 10 trials.

\item \textbf{Handover \& Place}: one arm picks up an object, transfers it to the other arm, and that arm places it into a container. The task is only solvable bimanually, since the object starts outside the placing arm's reachable workspace. Four stages: pick up the object, approach the handover configuration, complete the handover, and place the object. 10 trials, split over two different objects.

\item \textbf{Wipe Object} --- pick up a cloth and wipe a target surface with it. Four stages: pick up the cloth, align it with the target surface, wipe part of the surface, and cover the full area. Wiping appears in our egocentric human data but in none of the robot pre-training corpora. 10 trials.
\end{itemize}

\section{Additional Results}
\label{app:additional_results}

\subsection{Should the Action Expert Learn from Human Actions?}
\label{app:human_action_mask}

EgoLAP applies the flow-matching loss to human and robot action chunks alike, so the continuous action expert is trained on human hand motion as well as on robot end-effector motion. This is not an obvious choice: human motion is faster, noisier, and kinematically unlike anything the robot can execute, and one might expect it to corrupt an expert whose only job is real-time control. We therefore train a matched variant that masks the action-expert loss on every human sample while keeping the human language actions and motion-level reasoning that supervise the backbone. The variant isolates a single question: is human data useful only as \emph{language} supervision for the VLM, or also as \emph{motion} supervision for the expert?

Masking the loss is the worse choice overall (Fig.~\ref{fig:human_mask_ablation}). The masked variant loses $4.4$ points of overall success, and the loss is concentrated in the language-conditioned tasks, where it falls from $26.0\%$ to $10.0\%$; it is roughly on par on picking and placing and slightly better on cloth folding, the one task family whose dynamics are least like anything in the human corpus. We read this as evidence that the two roles of human data are not separable. Once the backbone is supervised with an action representation that already transfers across embodiments, the features the expert consumes are embodiment-agnostic enough that additional, noisier human motion acts as useful augmentation rather than as distribution shift.

\begin{figure}[H]
  \centering
  \includegraphics[width=\columnwidth]{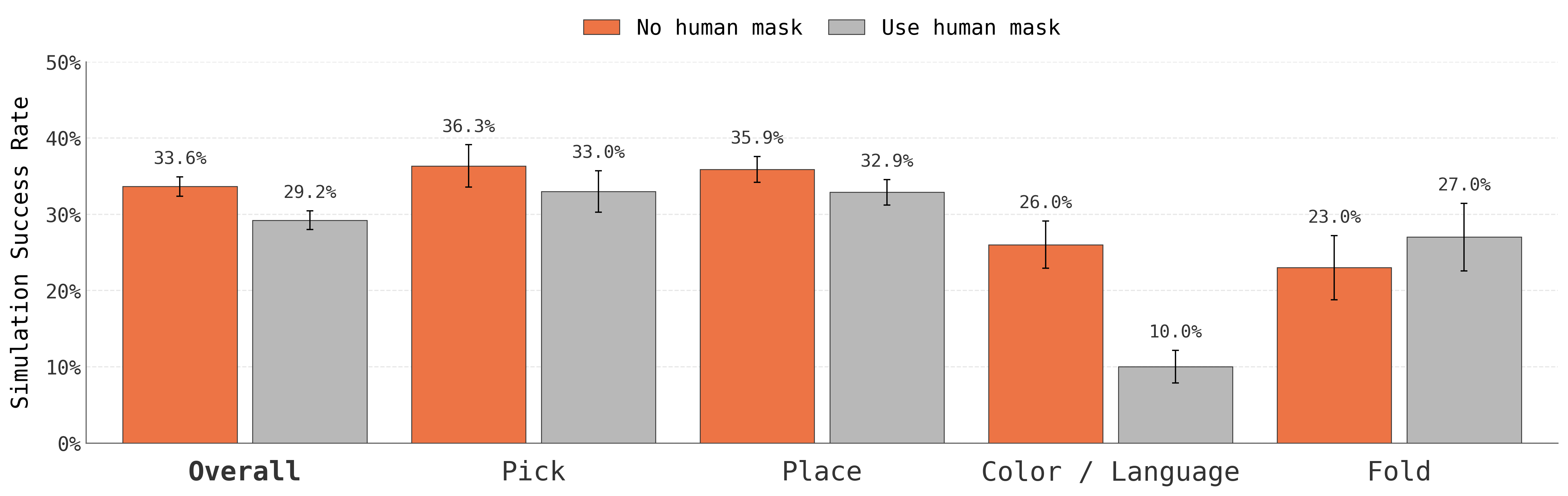}
  \caption{\textbf{Masking the action-expert loss on human samples hurts.} Simulation success over the 14-task suite for EgoLAP (``no human mask'', the main model) and for a matched variant that withholds the flow-matching loss on human samples (``use human mask''). Both variants receive identical language action and motion-level reasoning supervision; they differ only in whether the continuous action expert is trained on human motion. We compare the 25k step checkpoint for both models. Error bars are one binomial standard error.}
  \label{fig:human_mask_ablation}
\end{figure}

\subsection{Per-Task Simulation Results}
\label{app:per_task_simulation}

Figure~\ref{fig:per_task_simulation} expands the aggregate simulation results in Fig.~\ref{fig:results}~(a) into the 14 individual tasks and additionally includes the EgoOAT reasoning and no-reasoning variants. The aggregate gains are broadly distributed rather than driven by one task: EgoLAP outperforms its no-reasoning counterpart on 13 of 14 tasks, with especially large gains on cluttered fruit picking, Lego placement, colored-Lego selection, and towel folding. The task-level view also reveals complementary strengths that are hidden by the group means. For example, ABC-VLA is strongest on placing fruit in a bin and on a tray, whereas EgoLAP leads on most of the remaining placement tasks. EgoOAT reaches $3.3\%$ mean success with reasoning and $2.7\%$ without; its clearest reasoning gain is on towel folding ($15.6\%$ versus $2.1\%$).

\begin{figure}[H]
  \centering
  \includegraphics[width=0.99\textwidth]{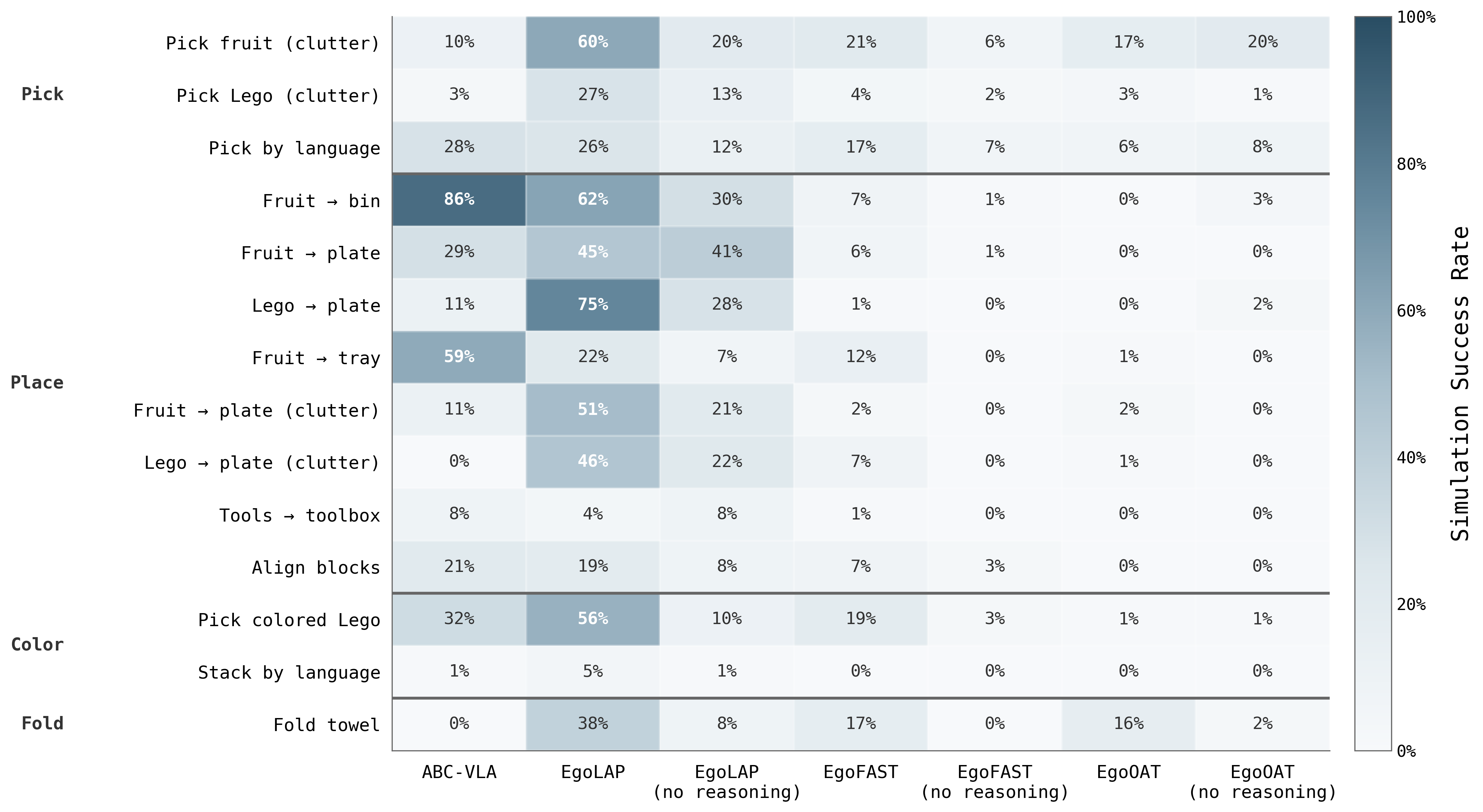}
  \caption{\textbf{Per-task zero-shot simulation success.} Rows expand the same 14-task suite summarized in Fig.~\ref{fig:results}~(a), grouped as Pick, Place, Color, and Fold. Columns show ABC-VLA and the reasoning and no-reasoning variants of EgoLAP, EgoFAST, and EgoOAT. Each cell reports the empirical task success rate, with darker shading indicating higher success.}
  \label{fig:per_task_simulation}
\end{figure}

\subsection{Do language actions benefit more from human data?}
\label{app:human_transfer_by_representation}

Higher performance under human--robot co-training could reflect a
stronger robot-only policy rather than a greater benefit from human
data. To distinguish these explanations, Table~\ref{tab:human_transfer}
compares robot-only and joint human--robot training for language
actions and FAST. We measure the human-data gain as the absolute
change in zero-shot simulation success when adding egocentric human
data.

\begin{table}[H]
    \centering
    \caption{\textbf{Language actions benefit more from human data
    under motion-level reasoning supervision.}
    Zero-shot success on the 14-task simulation suite.
    Human-data gains are absolute percentage-point differences.
    Reasoning denotes motion-level reasoning supervision.}
    \vspace{5pt}
    \label{tab:human_transfer}
    \begin{tabular}{llrrr}
        \toprule
        Representation & Reasoning & Robot only & Robot + human
        & Human-data gain \\
        \midrule
        Language & No  & 7.2\%  & 16.4\% & +9.2 pp \\
        Language & Yes & 18.4\% & 38.3\% & +19.9 pp \\
        FAST     & Yes & 1.4\%  & 8.7\%  & +7.3 pp \\
        \bottomrule
    \end{tabular}
\end{table}

With motion-level reasoning, adding human data improves language actions from 18.4\% to 38.3\%, a 19.9-point gain, whereas FAST improves from 1.4\% to 8.7\%, a 7.3-point gain. Thus, although language actions already provide a stronger robot-only baseline, their advantage widens under human--robot co-training: the gap over FAST increases from 17.0 to 29.6 points. This 12.6-point difference in human-data gains supports the conclusion that the advantage of language actions is not explained solely by stronger robot-only performance.

Within the language-action policies, human data also provides a
larger improvement with reasoning than without it
(19.9 versus 9.2 points), consistent with the interaction observed
in Section~\ref{subsec:reasoning_synergy} (Table~\ref{tab:robot_only_reasoning}). Together, these results support the effectiveness of pairing language actions with motion-level reasoning for
human-to-robot transfer under our training recipe.

\section{Generated Motion-Level Reasoning Examples and Prompt}
\label{app:reasoning_examples}

Following the annotation procedure in Appendix~\ref{app:reasoning_annotation}, we randomly show some motion-level reasoning labels from robot trajectories in DROID and egocentric human trajectories in MECKA, paired with the corresponding language actions, which form the complete textual supervision targets of the backbone. Each card places the unprocessed model observation on the left and the reasoning and language action on the right. Task headings are lightly normalized for readability.

\newcommand{\reasoningexamplecard}[5]{%
\begin{tcolorbox}[
  enhanced,
  colback=white,
  colframe=black!18,
  boxrule=0.55pt,
  arc=2mm,
  left=2mm,right=2mm,top=2mm,bottom=2mm,
  before skip=7pt,after skip=7pt,
  shadow={0.6mm}{-0.6mm}{0mm}{black!10}]
\begin{minipage}[t]{0.30\linewidth}
  \vspace{0pt}
  #2
\end{minipage}\hfill
\begin{minipage}[t]{0.66\linewidth}
  \vspace{0pt}
  {\scriptsize\bfseries\color{gtcolor}#1}\par
  \vspace{1pt}
  {\small\bfseries #3}\par
  \vspace{3pt}
  {\scriptsize\bfseries Motion-level reasoning}\par
  {\footnotesize #4}\par
  \vspace{3pt}
  {\scriptsize\bfseries Language action}\par
  {\scriptsize #5}
\end{minipage}
\end{tcolorbox}}

\newcommand{\reasoningexample}[5]{%
  \reasoningexamplecard{#1}{\includegraphics[width=\linewidth]{#2}}{#3}{#4}{#5}}

\newcommand{\reasoningexamplewrist}[6]{%
  \reasoningexamplecard{#1}{%
    {\centering\tiny\bfseries Primary\par}
    \includegraphics[width=\linewidth]{#2}\par
    \vspace{1mm}
    {\centering\tiny\bfseries Wrist\par}
    \includegraphics[width=\linewidth]{#3}
  }{#4}{#5}{#6}}

\subsection{DROID and MECKA Examples}
\label{app:droid_reasoning_examples}
\label{app:mecka_reasoning_examples}

\reasoningexamplewrist
  {DROID}
  {reasoning_examples/droid_08.jpg}
  {reasoning_examples/droid_08_wrist.jpg}
  {Push the pump of the blue bottle down}
  {The gripper has successfully grasped the pump nozzle. The robot now needs to push the pump down to fulfill the task instruction. The proposed motion moves the arm forward and down to begin depressing the pump mechanism. This downward motion must be controlled to avoid tipping the bottle.}
  {Left arm: move forward 1 cm, move down 2 cm, close gripper.}

\reasoningexamplewrist
  {DROID}
  {reasoning_examples/droid_02.jpg}
  {reasoning_examples/droid_02_wrist.jpg}
  {Hang the wire on the cabinet door}
  {With the hanger secured, the robot must lift it off the countertop to transport it to the cabinet. The next motion should lift the gripper upward and backward to clear the countertop and any obstacles. This begins the transport phase of the task.}
  {Left arm: move back 7 cm, move up 8 cm, move right 3 cm, tilt right 20 degrees, tilt forward 20 degrees, close gripper.}

\reasoningexamplewrist
  {DROID}
  {reasoning_examples/droid_15.jpg}
  {reasoning_examples/droid_15_wrist.jpg}
  {Put the sunglasses on the tray}
  {The gripper jaws are perfectly positioned around the sunglasses. The robot should move down slightly to ensure the jaws are fully surrounding the target, then close the gripper to secure the sunglasses. This action will establish a firm hold on the object.}
  {Left arm: move forward 2 cm, move down 4 cm, move right 2 cm, tilt back 10 degrees, rotate clockwise 10 degrees, close gripper.}

\reasoningexamplewrist
  {DROID}
  {reasoning_examples/droid_14.jpg}
  {reasoning_examples/droid_14_wrist.jpg}
  {Put the stick of glue in the mug}
  {The glue stick is now securely grasped. The robot needs to lift it to clear the table and the rim of the cup. The next motion should raise the gripper and move it toward the left where the cup is located.}
  {Right arm: move forward 1 cm, move up 14 cm, move left 25 cm, tilt right 10 degrees, tilt forward 10 degrees, rotate counterclockwise 20 degrees, close gripper.}

\reasoningexamplewrist
  {DROID}
  {reasoning_examples/droid_16.jpg}
  {reasoning_examples/droid_16_wrist.jpg}
  {Put the white object on the top shelf inside the grey toolbox}
  {The robot has successfully grasped and lifted the white USB adapter cable. The next step is to transport it toward the grey storage bin. The next motion should move the gripper forward and left while rotating to align with the bin.}
  {Left arm: move forward 4 cm, move left 8 cm, tilt forward 10 degrees, rotate clockwise 30 degrees, close gripper.}

\reasoningexample
  {MECKA}
  {reasoning_examples/mecka_v16_19.jpg}
  {Reposition the grip and continue brushing the metal component}
  {The left hand rests open across the metal disk, shifting slightly forward to maintain stability. Simultaneously, the right hand pulls the brushing tool upward and toward the torso, lifting the brush stroke clear to reposition for the next motion.}
  {Left arm: move forward 1 cm, move right 1 cm, move down 1 cm, tilt left 5 degrees, rotate counterclockwise 5 degrees, open gripper. Right arm: move backward 5 cm, move up 2 cm, tilt left 5 degrees, tilt backward 15 degrees, close gripper.}

\reasoningexample
  {MECKA}
  {reasoning_examples/mecka_2.png}
  {Insert small components into electronic device}
  {The left hand is directly above the open blue container. It should descend into the blue bin to target a small component. Translating down and forward aligns the fingers inside the container. The right hand should maintain its grip on the fixture.}
  {Left arm: move forward 6 cm, move down 11 cm, move right 12 cm, tilt right 10 degrees, tilt back 10 degrees, rotate clockwise 70 degrees, open gripper. Right arm: move back 4 cm, tilt left 10 degrees, tilt forward 10 degrees, rotate clockwise 20 degrees, close gripper.}

\reasoningexample
  {MECKA}
  {reasoning_examples/mecka_3.png}
  {Fill bottle with liquid from syringe}
  {The left hand supports the bottle firmly on the mat. The right hand shifts forward, up, and right while opening fingers to disengage from the plunger. This finishes the current stroke and allows resetting.}
  {Left arm: move back 2 cm, rotate counterclockwise 10 degrees, close gripper. Right arm: move forward 1 cm, move up 1 cm, move right 2 cm, tilt left 10 degrees, rotate clockwise 10 degrees, open gripper.}

\reasoningexample
  {MECKA}
  {reasoning_examples/mecka_4.png}
  {Cut red fabric into strips with scissors}
  {The left hand advances the held fabric over the table pile. The right hand keeps the scissors ready with fingers open. This repositions the active cutting zone conveniently.}
  {Left arm: move forward 5 cm, move down 1 cm, tilt right 10 degrees, tilt forward 10 degrees, rotate clockwise 10 degrees, close gripper. Right arm: move forward 4 cm, tilt forward 10 degrees, rotate counterclockwise 10 degrees, open gripper.}

\reasoningexample
  {MECKA}
  {reasoning_examples/mecka_5.png}
  {Open the bottle}
  {The right hand advances, lowers, closes fingers and rotates clockwise, securing a new contact point. The left hand shifts back and up slightly. This repositions the right hand for secondary stabilization. Closing the fingers provides firm support.}
  {Left arm: move back 1 cm, move up 2 cm, tilt right 10 degrees, tilt forward 10 degrees, rotate clockwise 10 degrees, open gripper. Right arm: move forward 2 cm, move down 2 cm, move right 5 cm, tilt right 50 degrees, rotate clockwise 40 degrees, close gripper.}

\subsection{Annotation Prompt}
\label{app:mecka_prompt}

We use separate prompts for MECKA and DROID. Below we retain the operative labeling instructions and show placeholders for per-step text, structured-action metadata, and images.

\begin{tcolorbox}[
  breakable,
  title={MECKA Gemini Prompt},
  colback=gtcolor!2,
  colframe=gtcolor!55,
  fonttitle=\bfseries,
  arc=2mm,
  boxrule=0.6pt]
\scriptsize
You label egocentric human bimanual hand-manipulation trajectories for training a vision-language action model. The model will be trained to predict a reasoning bridge first, then the provided language action. Generate concise, observable, action-grounded rationales.

These are MECKA first-person human manipulation videos. The primary image is the head-mounted egocentric camera view; there is no wrist camera. The action formatter is shared with robot data and may say ``Left arm,'' ``Right arm,'' and ``gripper.'' Interpret these as the human left and right hands or palms and the corresponding fingers. In generated reasoning, prefer human terms such as left hand, right hand, fingers, grasp, pinch, contact, support, and release.

A MECKA trajectory may be one subsequence cut from a longer high-level recording. Unless explicit stitched context is provided, treat the supplied instruction as the scope of the visible segment and do not invent unseen earlier or later subtasks. Build a concise trajectory summary, instruction-target checklist, and ordered subtask plan that cover all requested objects, receptacles, quantities, and spatial constraints. Mark requested but unseen entities as unresolved rather than substituting a different object.

For each requested step, use the current observations, task instruction, proposed language action, and compact past-action history when supplied. The requested steps are adaptive trajectory segments and may end early at a persistent finger-state change or major motion transition. Treat each image as the state before its segment action. Do not backdate the expected result of the current action or use later images as evidence that it succeeded.

Interpret finger opening and closing separately for the two hands. A commanded finger state does not by itself prove a grasp, release, or object support. Keep each hand's contact and role distinct, ground grasp or release claims in visible evidence, and respect the stated order between arm motion and a finger transition. Cross-check every directional word in the rationale against the supplied action; do not describe upward motion as lowering or downward motion as lifting.

The reasoning must connect visible task progress, the active hand--object relation, and the physical purpose of the next motion without copying action magnitudes or merely paraphrasing the command. State concrete relations to the target, receptacle, support surface, opening, rim, or obstacle. Explain relevant contact, support, stability, alignment, grasp choice, and clearance. If visibility is ambiguous, describe only what is supported by the image.

Return a trajectory summary, checklist, subtask plan, instruction paraphrases, and one label per requested step. Each step contains completed, current, and next subtasks; observation--action grounding; the main reasoning; a visual-relation sentence; a motion-rationale sentence; and an alternative-motion diagnostic. Return strict JSON only.

\medskip
\noindent\textbf{Trajectory and per-step input template}

\noindent{\ttfamily
Task instruction: [TASK INSTRUCTION]\par
Number of requested steps: [N]\par
\medskip
Step [STEP INDEX]: filtered trajectory index [INDEX], original frame [FRAME], segment frames [START]--[END] ([TERMINATION REASON]), duration [SECONDS]s, action-summary horizon [SECONDS]s.\par
Target language action in [ACTION FRAME]: [LANGUAGE ACTION]\par
Past 5s action history: [OPTIONAL HISTORY ITEMS]\par
\medskip
Structured-action metadata:\par
\{\par
\quad ``step\_index'': [STEP INDEX],\par
\quad ``trajectory\_step\_index'': [TRAJECTORY STEP INDEX],\par
\quad ``segment\_start'': [START], ``segment\_end'': [END],\par
\quad ``action\_frame'': ``egocentric frame'',\par
\quad ``language\_action'': ``[LANGUAGE ACTION]'',\par
\quad ``gripper\_open\_signal'': [LEFT/RIGHT FINGER STATE],\par
\quad ``history\_5s'': [OPTIONAL COMPACT ACTION HISTORY]\par
\}\par
\medskip
Primary camera image follows.\par
[[PRIMARY EGOCENTRIC IMAGE]]\par
\medskip
[REPEAT TEXT, PAYLOAD, AND IMAGE FOR EACH REQUESTED STEP]
}
\end{tcolorbox}

\begin{tcolorbox}[
  breakable,
  title={DROID Gemini Prompt},
  colback=bccolor!2,
  colframe=bccolor!55,
  fonttitle=\bfseries,
  arc=2mm,
  boxrule=0.6pt]
\scriptsize
You label robot manipulation trajectories for training a vision-language action model. The model will be trained to predict a reasoning bridge first, then the provided language action. Generate concise, observable, action-grounded rationales.

Use the task instruction and the temporally ordered primary- and wrist-camera observations to build a concise trajectory summary, instruction-target checklist, and ordered subtask plan. The plan must reflect the visible execution rather than a default grasp-and-drop script. Preserve every requested object, receptacle, quantity, color, order, and spatial constraint; mark a requested but unseen entity as unresolved rather than replacing or dropping it.

For each step, use the current observations and proposed target language action. Steps are adaptive trajectory segments and may end early at a persistent gripper-state change or a major motion transition. The prompt separates the current observation state from the future segment command. Describe the visible scene from the images and current sensor state, and use the future command only to explain what should happen next. Never describe a future grasp, release, contact, or placement as already complete.

Treat the raw gripper-open signal as an auxiliary cue; visible jaw contact and object support remain authoritative. An open or closed command is not by itself proof that an object is held or released. Respect whether the gripper transition occurs before, after, or concurrently with arm motion, and keep current-state wording distinct from the future transition. For each step, audit that the rationale agrees with the commanded translation, rotation, and gripper behavior. In particular, never reverse the commanded vertical direction in the explanation.

Use compact past-action history only to establish relevant progress; do not let history override the current images. Completed subtasks must precede the current step, the current subtask must name the phase advanced by the present low-level action, and the next subtask is only a preview. Do not claim that the whole task is complete while later object-directed actions remain.

The reasoning must connect visible progress, the gripper--object geometry, and the physical purpose of the next motion without copying action magnitudes or reciting the command. State concrete relations to the target, receptacle, support surface, opening, rim, table, or obstacle. Explain relevant contact, support, stability, alignment, grasp choice, and clearance. Prefer conservative descriptions when the primary and wrist views do not resolve a state.

Return a trajectory summary, checklist, subtask plan, instruction paraphrases, and one label per requested step. Each step contains completed, current, and next subtasks; observation--action grounding; the main reasoning; a visual-relation sentence; a motion-rationale sentence; and an alternative-motion diagnostic. Return strict JSON only.

\medskip
\noindent\textbf{Trajectory and per-step input template}

\noindent{\ttfamily
Task instruction: [TASK INSTRUCTION]\par
Number of requested steps: [N]\par
\medskip
Step [STEP INDEX]: filtered trajectory index [INDEX], original frame [FRAME], segment frames [START]--[END] ([TERMINATION REASON]), horizon [SECONDS]s.\par
Current observation state: [GRIPPER SENSOR STATE AND TRANSITION TIMING].\par
Future segment command in [ACTION FRAME]: [LANGUAGE ACTION]\par
Ordered action phases: [MOTION/GRIPPER PHASE SUMMARY]\par
Past 5s action history: [OPTIONAL HISTORY ITEMS]\par
\medskip
Structured-action metadata:\par
\{\par
\quad ``step\_index'': [STEP INDEX],\par
\quad ``trajectory\_step\_index'': [TRAJECTORY STEP INDEX],\par
\quad ``segment\_start'': [START], ``segment\_end'': [END],\par
\quad ``action\_frame'': ``egocentric frame'',\par
\quad ``current\_observation\_state'': ``[STATE SUMMARY]'',\par
\quad ``future\_segment\_command'': ``[LANGUAGE ACTION]'',\par
\quad ``gripper\_transition\_timing'': ``[START/END/CONCURRENT/NONE]'',\par
\quad ``history\_5s'': [OPTIONAL COMPACT ACTION HISTORY]\par
\}\par
\medskip
Primary camera image follows, then wrist camera image follows.\par
[[PRIMARY CAMERA IMAGE]]\par
[[WRIST CAMERA IMAGE]]\par
\medskip
[REPEAT TEXT, PAYLOAD, AND IMAGES FOR EACH REQUESTED STEP]
}
\end{tcolorbox}

}

\end{document}